\documentclass[acmsmall,screen]{acmart} 
\AtBeginDocument{%
  }

\setcopyright{acmlicensed}
\copyrightyear{2018}
\acmYear{2018}
\acmDOI{XXXXXXX.XXXXXXX}

\acmJournal{JACM}
\acmVolume{37}
\acmNumber{4}
\acmArticle{111}
\acmMonth{8}

\usepackage{amsmath}
\usepackage{multirow}
\usepackage{latexsym}
\usepackage{booktabs}
\usepackage{graphicx}
\usepackage{algorithmic}
\usepackage{algorithm}
\usepackage{array}
\usepackage[caption=false,font=normalsize,labelfont=sf,textfont=sf]{subfig}
\usepackage{textcomp}
\usepackage{stfloats}
\usepackage{url}
\usepackage{verbatim}
\usepackage{makecell}
\usepackage[table]{xcolor}
\usepackage[edges]{forest}
\usepackage{xcolor}
\usepackage{bm}
\colorlet{mypurple}{violet!30}
\colorlet{mygreen}{green!30}
\usepackage{wrapfig}
\usepackage{CJK}
\usepackage{tabularx}
\usepackage{hyperref}
\usepackage{tablefootnote}

\newcommand{\eg}{e.g.,~}
\newcommand{\etc}{etc.~}

\usepackage{microtype}

\begin{document}
\title{On The Role of Pretrained Language Models in General-Purpose Text Embeddings: A Survey}

\author{Meishan Zhang}
\email{mason.zms@gmail.com}
\orcid{0000-0001-6335-1340}
\affiliation{%
  \institution{Harbin Institute of Technology (Shenzhen)}
  \city{Shenzhen}
  \country{China}
}

\author{Xin Zhang}
\email{izhx404@gmail.com}
\orcid{0000-0002-2550-3056}
\affiliation{%
  \institution{Harbin Institute of Technology (Shenzhen)}
  \city{Shenzhen}
  \country{China}
}

\author{Xinping Zhao}
\email{zhaoxinping@stu.hit.edu.cn}
\orcid{0000-0001-6387-1442}
\affiliation{%
  \institution{Harbin Institute of Technology (Shenzhen)}
  \city{Shenzhen}
  \country{China}
}

\author{Shouzheng Huang}
\email{huangshouzheng@stu.hit.edu.cn}
\orcid{0009-0005-0169-9401}
\affiliation{%
  \institution{Harbin Institute of Technology (Shenzhen)}
  \city{Shenzhen}
  \country{China}
}

\author{Baotian Hu}
\email{hubaotian@hit.edu.cn}
\orcid{0009-0001-3210-167X}
\affiliation{%
  \institution{Harbin Institute of Technology (Shenzhen)}
  \city{Shenzhen}
  \country{China}
}

\author{Min Zhang}
\email{zhangmin2021@hit.edu.cn}
\orcid{0000-0002-3895-5510}
\affiliation{%
  \institution{Harbin Institute of Technology (Shenzhen)}
  \city{Shenzhen}
  \country{China}
}

\renewcommand{\shortauthors}{Zhang et al.}

\begin{abstract}
Text embeddings have attracted growing interest due to their effectiveness across a wide range of natural language processing (NLP) tasks, including retrieval, classification, clustering, bitext mining, and summarization. 
With the emergence of pretrained language models (PLMs), general-purpose text embeddings (GPTE) have gained significant traction for their ability to produce rich, transferable representations. 
The general architecture of GPTE typically leverages PLMs to derive dense text representations, which are then optimized through contrastive learning on large-scale pairwise datasets. 
In this survey, we provide a comprehensive overview of GPTE in the era of PLMs, focusing on the roles PLMs play in driving its development. 
We first examine the fundamental architecture and describe the basic roles of PLMs in GPTE, i.e., embedding extraction, expressivity enhancement, training strategies,  learning objectives, and data construction. 
We then describe advanced roles enabled by PLMs, including multilingual support, multimodal integration, code understanding, and scenario-specific adaptation. 
Finally, we highlight potential future research directions that move beyond traditional improvement goals, including ranking integration, safety considerations, bias mitigation, structural information incorporation, and the cognitive extension of embeddings. 
This survey aims to serve as a valuable reference for both newcomers and established researchers seeking to understand the current state and future potential of GPTE.
\footnote{If you have any suggestions for improvement, please contact \url{mason.zms@gmail.com} and \url{hubaotian@hit.edu.cn} without hesitation.}.
\end{abstract}



\received{20 February 2007}
\received[revised]{12 March 2009}
\received[accepted]{5 June 2009}

\maketitle

\begin{CJK}{UTF8}{gbsn}

\section{Introduction}
Text embeddings are continuous, dense, and fixed-size vectors used to represent discrete, variable-length texts, enabling large-scale automated computation and analysis of textual data. 
It is a critical technology widely applied in natural language processing (NLP) \cite{muennighoff-etal-2023-mteb} and information retrieval (IR) \cite{huang2020embedding}. 
Embeddings can serve as high-level features for various text classification tasks,  
and can also be exploited to calculate distances between texts, which is essential for clustering with algorithms like K-Nearest Neighbors (KNN) \cite{grootendorst2022bertopic}. 
By capturing the semantic meaning of texts in a vector space, embeddings allow for direct computation of similarity measures (e.g., cosine similarity), supporting tasks such as retrieval, reranking, semantic textual similarity (STS), and natural Language inference (NLI). 
Furthermore, embedding-based retrievers are a key component of retrieval-augmented generation (RAG) \cite{gao2023retrieval,DBLP:conf/naacl/ZhaoZSHLLHZ25,DBLP:conf/nips/LewisPPPKGKLYR020}, enabling generative large language models (LLMs) to access up-to-date external knowledge without altering their parameters.

Early works on embeddings include Latent Semantic Indexing (LSA) \cite{deerwester1990indexing} and Latent Dirichlet Allocation (LDA) \cite{blei2003latent}. 
With the advent of deep learning and word embeddings \cite{pennington2014glove,mikolov2013linguistic}, various neural network models such as direct pooling, convolutional neural networks (CNNs), and recurrent neural networks (RNNs), have been developed to generate supervised text representations in the form of embeddings \cite{le2014distributed,kiros2015skip,hill2016learning,kalchbrenner2014convolutional}. 
These embeddings are typically derived as by-products, as the primary goal of these models is aimed at a specific downstream task \cite{conneau2017supervised}.
In parallel, learning general-purpose text embeddings (GPTE) in an unsupervised, self-supervised, or multiple-task-supervision manner has attracted significant attention \cite{cer2018universal,kiros2015skip,DBLP:conf/emnlp/YangYCLD21}. 
Such embeddings can be directly applied to a wide range of tasks in a universal manner, making this line of work a prominent direction in text embedding research.
Currently, pretrained language models (PLMs) have significantly transformed natural language processing tasks \cite{devlin-etal-2019-bert}, without any exception for text embeddings as well.
For instance, we can obtain text embeddings by feeding a text into BERT-alike models directly, using the vectorial representation of $\langle CLS \rangle$ as the full-text embedding.
The derived embeddings were unprecedentedly powerful in strength of general purpose and can be further optimized through supervised fine-tuning (SFT) \cite{howard2018universal}, following the pretrain-then-finetune paradigm.  
As PLMs continue to evolve, from encoder-based \cite{devlin-etal-2019-bert} to encoder-decoder architectures \cite{raffel2020exploring},  and more recently to decoder-only large language models (LLMs) \cite{brown2020language}, the development of GPTE has also progressed rapidly.
According to the widely recognized MTEB benchmark website\footnote{\url{https://huggingface.co/spaces/mteb/leaderboard}}, at least 150 text embedding models have been released in recent years, reflecting the growing interest and progress in this area. 
Text embeddings are conceptually straightforward, and their overall architecture has remained largely consistent since the introduction of GPTE. 
Typically, a deep neural network is first initialized to compose text representations from word embeddings \cite{DBLP:conf/naacl/PetersNIGCLZ18}. 
Then, unsupervised, self-supervised, or weakly supervised objectives are used to train the network parameters, resulting in GPTE \cite{DBLP:conf/emnlp/GaoYC21}. 
Finally, task-specific SFT is applied to align the GPTE with the downstream tasks \cite{DBLP:conf/emnlp/ReimersG19},
which is an almost trivial step within the PLM era.   
Based on the observation, a natural question arises: \textit{What role do PLMs play in advancing GPTE models?}
The answer lies in examining two distinct angles: (1) the fundamental role of PLMs in enhancing the capability of GPTE, and (2) their advanced role in expanding the scope of GPTE applications.

In this paper, we present a comprehensive survey investigating the role of PLMs in GPTE.
We systematically review the representative GPTE models built on PLMs, which aim to offer a clear and accessible roadmap for understanding recent developments in this field.
Section \ref{sec:background} provides a brief background on GPTE, describing the basic concepts of GPTE and outlining the general learning architecture to establish a foundation for examining PLM contributions.
Subsequently, we categorize key works of exploring PLMs for text embedding, and describe the successive methods within each aspect in detail, highlighting the related research efforts in terms of roles in Section~\ref{sec:role}-\ref{sec:advance}.
Based on our observations and insights, we propose several promising future directions in Section \ref{sec:future}, and finally conclude this paper in Section \ref{sec:conclu}. 
The overall taxonomy of PLMs’ Roles in GPTE is illustrated in Figure~\ref{fig:taxo}.

Several existing surveys are closely related to text embeddings. 
Among them, surveys on sentence representations are particularly relevant \cite{ramesh-kashyap-etal-2024-comprehensive,DBLP:journals/csur/LiZM23}, with the main distinction lying in target granularity and application scope.
Sentence representations typically focus on capturing sentence-level semantics, while text embeddings are designed for a broader range of texts and applications.
Surveys on the use of PLMs in IR also touch on text embeddings, though their emphasis remains on IR-specific tasks \cite{zhao2024dense,wang2024utilizing,guo2022semantic}. 
The most directly related works include reviews of universal text embeddings \cite{cao2024recent} and LLM-empowered text embeddings \cite{nie2024text}.
The former primarily lists top-performing models on the MTEB benchmark, whereas the latter offers an overview of LLM-based methods across several dimensions, which differ significantly from our approach.
In contrast, our survey adopts a unified, architecture-driven perspective, tracing the evolution of PLMs within this framework. 
In particular, we give equal emphasis to GPTE models based on BERT-like architectures since they are still important in real-application settings.

\tikzstyle{my-box}=[
    rectangle,
    draw=gray!50,
    rounded corners,
    text opacity=1,
    minimum height=2em,
    minimum width=5em,
    inner sep=3pt,
    inner ysep=6pt,
    align=center,
    fill opacity=0.15,
    line width=0.5pt,
]

\tikzstyle{leaf}=[my-box, minimum height=2em,
    fill=gray!5, text=black, align=left, font=\normalsize,
    inner xsep=3pt,
    inner ysep=6pt,
    line width=0.5pt,
]

\definecolor{c1}{RGB}{102,178,255}
\definecolor{c2}{RGB}{255,153,153}
\definecolor{c3}{RGB}{255,204,102}
\definecolor{c4}{RGB}{153,221,153}
\definecolor{c5}{RGB}{204,179,255}
\definecolor{c7}{RGB}{153,221,214}
\definecolor{c8}{RGB}{221,160,221}
\definecolor{c9}{RGB}{255,179,207}

\begin{figure*}[!tp]
    \centering
    \resizebox{\textwidth}{!}{
        \begin{forest}
            forked edges,
            for tree={
                grow=east,
                reversed=true,
                anchor=base west,
                parent anchor=east,
                child anchor=west,
                base=center,
                font=\large,
                rectangle,
                draw=gray,
                rounded corners,
                align=left,
                text centered,
                minimum width=4em,
                edge+={darkgray, line width=0.5mm},
                s sep=3pt,
                inner xsep=2pt,
                inner ysep=3pt,
                line width=0.8pt,
                ver/.style={rotate=90, child anchor=north, parent anchor=south, anchor=center},
            },
            where level=1{text width=10em,font=\normalsize,}{},
            where level=2{text width=14em,font=\normalsize,}{},
            where level=3{text width=10em,font=\normalsize,}{},
            where level=4{text width=38em,font=\normalsize,}{},
            where level=5{text width=10em,font=\normalsize,}{},
            [
                {\Large \textbf{On The Role of PLMs in GPTE}}, ver, line width=0.7mm
                [
                    {\large \shortstack{\textbf{Background~(\S\ref{sec:background})}}}, fill=c1!60, draw=c1, line width=0.5mm
                    [
                        \shortstack{\textbf{Concept of} \\ \textbf{Text Embeddings~(\S\ref{sec:concepts})}}, 
                        fill=c1!60, draw=c1, line width=0.5mm, edge={c1}
                        [
                        A Vector Space Model for Automatic Indexing~\cite{DBLP:journals/cacm/SaltonWY75}; Downstream NLP tasks~\cite{da2023text,wehrli-etal-2023-german,DBLP:conf/coling/YanoFFTW24}, leaf, text width=38em, draw=c1, line width=0.7mm, edge={c1}
                        ]
                    ]
                    [
                        \shortstack{\textbf{Applications of} \\ \textbf{Text Embeddings~(\S\ref{sec:applications})}}, 
                        fill=c1!60, draw=c1, line width=0.5mm, edge={c1}
                        [
                            \shortstack{\textbf{Semantic}  \\ \textbf{Similarity~(\S\ref{sec:ss})}}, fill=c1!60, draw=c1, line width=0.5mm, edge={c1}
                            [
                                STS~\cite{DBLP:conf/semeval/CerDALS17,chandrasekaran2021evolution,DBLP:conf/emnlp/HuCZ15}; NLI~\cite{bowman2015large,conneau2017supervised,DBLP:conf/emnlp/KoreedaM21}; Clustering~\cite{keraghel2024beyond,petukhova2024text,DBLP:journals/corr/abs-2104-07081}, leaf, text width=26.5em, draw=c1, line width=0.7mm, edge={c1}
                            ]
                        ]
                        [
                            \shortstack{\textbf{Semantic}  \\ \textbf{Relevance~(\S\ref{sec:sr})}}, fill=c1!60, draw=c1, line width=0.5mm, edge={c1}
                            [
                                IR \cite{manning2009introduction,guo2022semantic}; QA \cite{DBLP:journals/tacl/KwiatkowskiPRCP19,karpukhin-etal-2020-dense}; RAG~\cite{gao2023retrieval,DBLP:conf/naacl/ZhaoZSHLLHZ25,DBLP:conf/nips/LewisPPPKGKLYR020}; LLM Memory\\~\cite{park2023generative,zhong2024memorybank}; Fact Verification~\cite{DBLP:conf/naacl/ThorneVCM18}, leaf, text width=26.5em, draw=c1, line width=0.7mm, edge={c1}
                            ]
                        ]
                        [
                            \shortstack{\textbf{Semantic}  \\ \textbf{Encoding~(\S\ref{sec:se})}}, fill=c1!60, draw=c1, line width=0.5mm, edge={c1}
                            [
                                Text Classifiers~\cite{da2023text}; Semantic Reasoning Systems~\cite{cheng2024xrag}; Image \\ Generators~\cite{saharia2022photorealistic}, leaf, text width=26.5em, draw=c1, line width=0.7mm, edge={c1}
                            ]
                        ]
                        [
                            \shortstack{\textbf{Hybrid}  \\ \textbf{Combination~(\S\ref{sec:hc})}}, fill=c1!60, draw=c1, line width=0.5mm, edge={c1}
                            [
                                xRAG~\cite{cheng2024xrag}; UniRAG~\cite{DBLP:conf/acl/LiHLZYYZS25}; Think-Then-Embed~\cite{cui2025think}, leaf, text width=26.5em, draw=c1, line width=0.7mm, edge={c1}
                            ]
                        ]
                    ]
                    [
                        \shortstack{\textbf{The General Architecture} \\ \textbf{of GPTE~(\S\ref{sec:common:architecture})}}, 
                        fill=c1!60, draw=c1, line width=0.5mm, edge={c1}
                        [
                        CL~\cite{DBLP:conf/coling/ChenZS022,DBLP:conf/naacl/NishikawaRYTE22,DBLP:conf/acl/KleinN22,DBLP:conf/coling/WuGSHWH22,DBLP:conf/acl/ZhangZWXLZ22}; Margin-based CL~\cite{DBLP:conf/acl/ZhangZWXLZ22}; Momentum Smoothing~\cite{DBLP:conf/coling/0002GZHWH22,DBLP:conf/icse/ShiWGDZHZS23}, leaf, text width=38em, draw=c1, line width=0.7mm, edge={c1}
                        ]
                    ]
                    [
                        \shortstack{\textbf{Training Dataset~(\S\ref{sec:training:data})}}, 
                        fill=c1!60, draw=c1, line width=0.5mm, edge={c1}
                        [
                            Retrieval~\cite{karpukhin-etal-2020-dense,li2023towards,li2024making}; Classification~\cite{kim-2014-convolutional,howard2018universal,da2023text}; STS~\cite{DBLP:conf/emnlp/ReimersG19,hill2016learning}; QA~\cite{shi2024replug,guu2020retrieval}, leaf, text width=38em, draw=c1, line width=0.7mm, edge={c1}
                        ]
                    ]
                    [
                        \shortstack{\textbf{GPTE Evaluation~(\S\ref{sec:gpte:eval})}}, 
                        fill=c1!60, draw=c1, line width=0.5mm, edge={c1}
                        [
                        MTEB~\cite{muennighoff-etal-2023-mteb}; MMTEB~\cite{enevoldsen2025mmteb}; C-MTEB~\cite{xiao2023c}; SentEval~\cite{DBLP:conf/lrec/ConneauK18}; USEB~\cite{DBLP:journals/corr/abs-2104-06979}; SciRepEval~\cite{DBLP:conf/emnlp/SinghDCDF23}, leaf, text width=38em, draw=c1, line width=0.7mm, edge={c1}
                        ]
                    ]
                ]
                [   
                    {\large \shortstack{\textbf{Basic Roles~(\S\ref{sec:role})}}}, fill=c9!60, draw=c9, line width=0.5mm
                    [
                        \shortstack{\textbf{Text Embedding} \\ \textbf{Acquisition~(\S\ref{sec:embedding:acquire})}}, align=center, fill=c9!60, draw=c9, line width=0.5mm, edge={c9}
                        [
                            Encoder-based PLMs~\cite{devlin-etal-2019-bert}; Decoder-based PLMs~\cite{radford2018improving}; Pooling~\cite{ni-etal-2022-sentence,lee2024nv,oh-etal-2022-dont,su2021whitening}, leaf, text width=38em, draw=c9, line width=0.7mm, edge={c9}
                        ]
                    ]
                    [
                        \shortstack{\textbf{Expressibility} \\ \textbf{Improvement~(\S\ref{sec:role:express})}}, align=center, fill=c9!60, draw=c9, line width=0.5mm, edge={c9}
                        [
                            \shortstack{\textbf{Long-Context}  \\ \textbf{Modeling~(\S\ref{sec:long:modeling})}}, fill=c9!60, draw=c9, line width=0.5mm, edge={c9}
                            [
                                Plug-and-Play Augmentation~\cite{chen2024bge,zhu2024longembed,sturua2024jina3};  Pre-Training \\ from Scratch~\cite{portes2024mosaicbert,gunther2023jina2,saad2024benchmarking,warner2024smarter}, leaf, text width=26.5em, draw=c9, line width=0.7mm, edge={c9}
                            ]
                        ]
                        [
                            \shortstack{\textbf{Prompt-Informed} \\ \textbf{Embedding~(\S\ref{sec:prompt:emb})}}, fill=c9!60, draw=c9, line width=0.5mm, edge={c9}
                            [
                                PromptBERT~\cite{jiang-etal-2022-promptbert}; PromCSE~\cite{DBLP:conf/emnlp/JiangZW22}; PCoTEOL~\cite{zhang2024simple}; Echo\\~\cite{springer2024repetition}; E5-mistral~\cite{wang2023improving}; GritLM~\cite{muennighoff2024generative}; GEM~\cite{zhang2025gem},leaf, text width=26.5em, draw=c9, line width=0.7mm, edge={c9}
                            ]
                        ]
                    ]
                    [
                         \shortstack{\textbf{Parameter} \\ \textbf{Optimization~(\S\ref{sec:role:opt})}}, fill=c9!60, draw=c9, line width=0.5mm, edge={c9}
                        [
                            \shortstack{\textbf{Multi-Stage} \\ \textbf{Training~(\S\ref{sec:multi:stage})}},
                            fill=c9!60, draw=c9, line width=0.5mm, edge={c9}
                            [
                               Improving the text embedding quality  progressively with \\ weakly-supervised data and high-quality supervised data\\~\cite{wang2022text,li2023towards,qwen3embedding,hu2025kalm,zhao2025kalmembeddingv2}, leaf, text width=26.5em, draw=c9, line width=0.7mm, edge={c9}
                            ]
                        ]
                        [
                            \shortstack{\textbf{Objectives Beyond} \\ \textbf{Contrastive Learning} \\ \textbf{(\S\ref{sec:obj:beyong:cl})}}, fill=c9!60, draw=c9, line width=0.5mm, edge={c9}
                            [
                                Joint Training with GPTE Objectives~\cite{DBLP:conf/emnlp/SchickS21a,DBLP:conf/naacl/ChuangDLZCS0YKG22,DBLP:conf/emnlp/WuGLHWH22,DBLP:conf/emnlp/WuZ22,muennighoff2024generative,behnamghader2024llm2vec}; \\ Matryoshka Representation Learning (MRL)~\cite{kusupati2022matryoshka}; CoSent~\cite{su2022cosent}; \\ Knowledge distillation~\cite{DBLP:conf/sigir/Cheng21,wang2022text,DBLP:conf/acl/SeonwooWSCLLXPO23,DBLP:conf/acl/LiuLWWWX0C023,chen2024bge,zhao2025kalmembeddingv2} \etc, leaf, text width=26.5em, draw=c9, line width=0.7mm, edge={c9}
                            ]
                        ]
                        [
                            \shortstack{\textbf{Batch Learning} \\ \textbf{(\S\ref{sec:batch:learn})}}, fill=c9!60, draw=c9, line width=0.5mm, edge={c9}
                            [
                                GradCache~\cite{gao-etal-2021-scaling}; GistEmbed~\cite{solatorio2024gistembed}; Activation Checkpointing\\~\cite{chen2016training}; Zero Redundancy Optimizer~\cite{rajbhandari2020zero}, leaf, text width=26.5em, draw=c9, line width=0.7mm, edge={c9}
                            ]
                        ]
                    ]
                    [
                         \shortstack{\textbf{Data Synthesis~(\S\ref{sec:role:data})}}
                         , fill=c9!60, draw=c9, line width=0.5mm, edge={c9}
                        [
                            \shortstack{\textbf{Training Data} \\ \textbf{Synthesis~(\S\ref{sec:train:synth})}},
                            fill=c9!60, draw=c9, line width=0.5mm, edge={c9}
                            [
                               Training Data Synthesis~\cite{lee2024gecko,lee2024nv,wang2023improving,hu2025kalm,qwen3embedding}; Persona-based \\ Synthesis~\cite{hu2025kalm,qwen3embedding,zhao2025kalmembeddingv2}; Semantically Similar Text Synthesis\\~\cite{DBLP:conf/emnlp/ZhangLH23,DBLP:conf/naacl/OuX24,DBLP:conf/naacl/ThirukovalluruWCLLJD24}; Negative Synthesis~\cite{DBLP:conf/emnlp/GaoYC21,kim2021self,zhang2022virtual,cao2022exploring,zhou2022debiased,zhang2022unsupervised},leaf, text width=26.5em, draw=c9, line width=0.7mm, edge={c9}
                            ]
                        ]
                        [
                            \shortstack{\textbf{Evolving Benchmark} \\ \textbf{(\S\ref{sec:evolv:bench})}}, fill=c9!60, draw=c9, line width=0.5mm, edge={c9}
                            [
                                AIR-bench~\cite{chen2024air}, leaf, text width=26.5em, draw=c9, line width=0.7mm, edge={c9}
                            ]
                        ]
                    ]
                    [
                         \shortstack{\textbf{The Choice of PLMs~(\S\ref{sec:role:choice})}}
                         , fill=c9!60, draw=c9, line width=0.5mm, edge={c9}
                        [
                            \shortstack{\textbf{Comparison of Model} \\ \textbf{Architectures~(\S\ref{sec:comp:arch})}},
                            fill=c9!60, draw=c9, line width=0.5mm, edge={c9}
                            [
                               Encoder-based GPTE~\cite{DBLP:conf/emnlp/ReimersG19,wang2022text,li2023towards,xiao2023c}; Decoder-based GPTE\\~\cite{zhang2023language,wang2023improving,zhang2023language,wang2023improving,lee2024nv,li2023towards},leaf, text width=26.5em, draw=c9, line width=0.7mm, edge={c9}
                            ]
                        ]
                        [
                            \shortstack{\textbf{Impact of Model} \\ \textbf{Scale~(\S\ref{sec:imp:scale})}}, fill=c9!60, draw=c9, line width=0.5mm, edge={c9}
                            [
                                Scaling Effect~(Figure~\ref{fig:trend_perf_emb}); MoE-based GPTE~\cite{muennighoff2024generative,nussbaum2025nomicmoe,DBLP:conf/iclr/LiZ25}, leaf, text width=26.5em, draw=c9, line width=0.7mm, edge={c9}
                            ]
                        ]
                    ]
                ]
                [
                    {\large \shortstack{\textbf{Advanced Roles~(\S\ref{sec:advance})}}},
                    fill=c4!60, draw=c4, line width=0.5mm
                    [
                        \shortstack{\textbf{Multilingualism~(\S\ref{sec:multilingual})}},
                        align=center, fill=c4!60, draw=c4, line width=0.5mm, edge={c4}
                        [
                            Multilingual GPTE Models~\cite{DBLP:journals/tmlr/IzacardCHRBJG22,DBLP:journals/tois/ZhangOML24,wang2024multilingual,DBLP:conf/emnlp/ZhangZLXDTLYXHZ24, qwen3embedding,DBLP:journals/corr/abs-2402-03216,zhao2025kalmembeddingv2,DBLP:conf/emnlp/Ni0LDAMZLHCY22,hu2025kalm,feng2022language,DBLP:conf/coling/YanoFFTW24}~(Table~\ref{tab:multiemb}); \\ Multilingual Training Data~\cite{DBLP:journals/corr/abs-2104-07081,sentence-transformers-reddit,DBLP:journals/corr/abs-2104-07081,sentence-transformers-reddit,wikidump,xue2021mt5,DBLP:conf/lrec/WenzekLCCGJG20,DBLP:conf/acl/MuennighoffWSRB23,DBLP:journals/corr/abs-2109-12870,DBLP:conf/acl/LewisORRS20,DBLP:journals/tacl/LongpreLD21}~(Table~\ref{tab:multidata}), leaf, text width=38em, draw=c4, line width=0.7mm, edge={c4}
                        ]
                    ]
                    [
                        \textbf{Multimodal~(\S\ref{sec:multimodal})},
                        fill=c4!60, draw=c4, line width=0.5mm, edge={c4}
                        [
                            Individual Encoder~\cite{DBLP:conf/icml/RadfordKHRGASAM21,DBLP:conf/icml/0001LXH22,DBLP:conf/icml/JiaYXCPPLSLD21,DBLP:conf/iccv/ZhaiM0B23,DBLP:journals/tmlr/YuWVYSW22,DBLP:journals/corr/abs-2502-14786,DBLP:conf/icml/0008LSH23}; Unified Encoder\\~\cite{DBLP:conf/iclr/JiangMYYZC25,meng2025vlm2vec,DBLP:journals/corr/abs-2504-17432,DBLP:journals/corr/abs-2503-04812,DBLP:journals/corr/abs-2505-11293,DBLP:journals/corr/abs-2505-19650}~(Table~\ref{tab:mmemb}); Multimodal Training Data~\cite{DBLP:journals/corr/abs-2412-14475,DBLP:journals/corr/abs-2412-16855,DBLP:journals/corr/abs-2502-08468}~(Table~\ref{tab:mmdata}), leaf, text width=38em, draw=c4, line width=0.7mm, edge={c4}
                        ]
                    ]
                    [
                        \shortstack{\textbf{Programming Languages} \\ \textbf{(\S\ref{sec:prog:lang})}},
                        fill=c4!60, draw=c4, line width=0.5mm, edge={c4}
                        [
                       Code Embeddig Models~\cite{DBLP:journals/corr/abs-2004-13214,DBLP:conf/emnlp/FengGTDFGS0LJZ20,DBLP:conf/icml/KanadeMBS20,DBLP:conf/iclr/GuoRLFT0ZDSFTDC21,wang2021syncobert,DBLP:conf/uai/JiangZLLL21,DBLP:conf/acl/GuoLDW0022,DBLP:conf/nips/LachauxRSL21,roziere2020unsupervised,DBLP:conf/naacl/AhmadCRC21,DBLP:conf/emnlp/0034WJH21,DBLP:journals/corr/abs-2105-08645}~(Table~\ref{tab:codeemb}); \\ Code Training Data~\cite{puri2codenet,DBLP:conf/iclr/SureshRXNMDJ25,lu1codexglue,mou2016convolutional,ahmad2021avatar,zhu2022multilingual,yan2023codetransocean,roziereleveraging,svajlenko2014towards,DBLP:conf/acl/HuangTSG0J0D20,iyer2018mapping,DBLP:conf/nips/HendrycksBKMAGB21}~(Table~\ref{tab:codedata}), leaf, text width=38em, draw=c4, line width=0.7mm, edge={c4}
                        ]
                    ]
                    [
                        \shortstack{\textbf{Adaptation for} \\ \textbf{Specific Scenarios~(\S\ref{sec:adapt:spec})}},
                        fill=c4!60, draw=c4, line width=0.5mm, edge={c4}
                        [
                           Instruction-following GPTE~\cite{li2024making,DBLP:conf/nips/Ouyang0JAWMZASR22,DBLP:conf/iclr/SanhWRBSACSRDBX22,DBLP:conf/acl/MishraKBH22,DBLP:conf/iclr/WeiBZGYLDDL22}; Language-specific GPTE~\cite{xiao2023c,DBLP:journals/corr/abs-2402-06617}; \\ Domain-specific Adaptation~\cite{DBLP:journals/bioinformatics/LeeYKKKSK20,DBLP:journals/corr/abs-1904-03323,DBLP:journals/corr/abs-2412-15258}, leaf, text width=38em, draw=c4, line width=0.7mm, edge={c4}
                        ]
                    ]
                ]
                [
                    {\large \shortstack{\textbf{Expected Role~(\S\ref{sec:future}})}},
                    fill=c5!60, draw=c5, line width=0.5mm
                    [
                        \shortstack{\textbf{Combination with} \\ \textbf{Text Ranking~(\S\ref{sec:comb:rank})}},
                        align=center, fill=c5!60, draw=c5, line width=0.5mm, edge={c5}
                        [
                            Jina-reranker-v3~\cite{DBLP:journals/corr/abs-2509-25085}; Qwen3-Reranker~\cite{qwen3embedding}, leaf, text width=38em, draw=c5, line width=0.7mm, edge={c5}
                        ]
                    ]
                    [
                        \shortstack{\textbf{Safety Considerations} \\ \textbf{(\S\ref{sec:safety:consi})}},
                        fill=c5!60, draw=c5, line width=0.5mm, edge={c5}
                        [
                            BadCSE~\cite{DBLP:journals/corr/abs-2210-11082}; GEIA~\cite{DBLP:conf/acl/0003XS23}; ALGEN~\cite{DBLP:journals/corr/abs-2502-11308}; Revealer~\cite{DBLP:journals/corr/abs-2209-10505}; TEIA~\cite{DBLP:conf/acl/HuangTHLL24}, leaf, text width=38em, draw=c5, line width=0.7mm, edge={c5}
                        ]
                    ]
                    [
                        \shortstack{\textbf{Bias of GPTE~(\S\ref{sec:bias:gpte})}},
                        fill=c5!60, draw=c5, line width=0.5mm, edge={c5}
                        [
                            Debiasing Word Embeddings~\cite{DBLP:conf/nips/BolukbasiCZSK16}; Task-diverse Pretraining Regimes~\cite{qwen3embedding}; Domain-\\adaptive Pretraining~\cite{DBLP:journals/corr/abs-2504-20595}; Multilingual Alignment and Language-specific Tuning\\\cite{DBLP:conf/emnlp/ZhangZLXDTLYXHZ24,DBLP:journals/corr/abs-2412-08802}; Modality-invariant Representation~\cite{DBLP:journals/corr/abs-2404-07983}, leaf, text width=38em, draw=c5, line width=0.7mm, edge={c5}
                        ]
                    ]
                    [
                        \shortstack{\textbf{Structure Information} \\ \textbf{(\S\ref{sec:struct:info})}},
                        fill=c5!60, draw=c5, line width=0.5mm, edge={c5}
                        [
                            Tables~\cite{DBLP:conf/acl/YinNYR20}; Codes~\cite{DBLP:conf/emnlp/FengGTDFGS0LJZ20}; Knowledge Graphs~\cite{DBLP:journals/tacl/WangGZZLLT21}, leaf, text width=38em, draw=c5, line width=0.7mm, edge={c5}
                        ]
                    ]
                    [
                        \shortstack{\textbf{Extending GPTE with} \\ \textbf{Reasoning~(\S\ref{sec:gpte:reason})}},
                        fill=c5!60, draw=c5, line width=0.5mm, edge={c5}
                        [
                            Think-Then-Embed~\cite{cui2025think}; O1-Embedder~\cite{DBLP:journals/corr/abs-2502-07555}, leaf, text width=38em, draw=c5, line width=0.7mm, edge={c5}
                        ]
                    ]
                ]
            ]
        \end{forest}
    }
    \caption{Taxonomy of PLMs' Roles in GPTE.}
    \vspace{-2pt}
    \label{fig:taxo}
\end{figure*}

\section{Background}\label{sec:background}

\subsection{Concept of Text Embeddings}\label{sec:concepts}

Text is a discrete and variable-length sequence of words, sentences, or paragraphs. 
To make it amenable to computation, we employ well-designed models to encode its underlying semantics into fixed-size vector representations (real-valued vector $\bm{e} \in \mathbb{R}^{d}$, $d$ is the dimension size), called text embeddings.
The embedding model can serve as the foundation of the vector space of texts~\cite{DBLP:journals/cacm/SaltonWY75}, 
enabling the calculation of quantified semantic similarities or distance measures, an essential capability for semantic understanding and IR. 
Additionally, the derived embeddings can be leveraged as high-level semantic features, making them adaptable for a wide range of downstream NLP tasks, such as classification, clustering, STS, question-answering, and so on~\cite{da2023text,wehrli-etal-2023-german,DBLP:conf/coling/YanoFFTW24}.
\begin{figure}
    \centering
    \includegraphics[width=.99\linewidth]{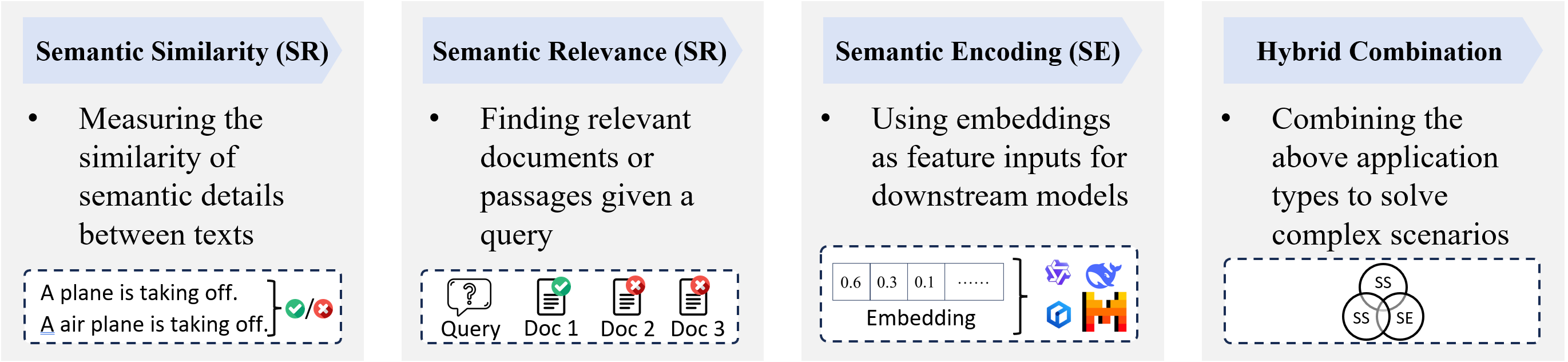}
    \caption{Four typical applications of text embedding.}
    \label{fig:application}
\end{figure}

\subsection{Applications of Text Embeddings}\label{sec:applications}
Text embedding applications can be broadly categorized into three types based on their primary purpose: semantic similarity, semantic relevance, and semantic encoding. 
The first two focus on bi-text semantic computations, where the last type involves representing individual texts as high-level features for downstream tasks. 
Semantic similarity typically refers to symmetric tasks where both texts are treated equally, while semantic relevance addresses asymmetric tasks where one text is semantically related to another. 
Additionally, several hybrid cases exist within text embedding applications.
Below, we first introduce the three basic types of applications, and then provide examples of hybrid applications, illustrated in Figure~\ref{fig:application}.

\subsubsection{Semantic Similarity (SS)}\label{sec:ss}
This type of application primarily includes STS~\cite{DBLP:conf/semeval/CerDALS17,chandrasekaran2021evolution}, NLI~\cite{bowman2015large,conneau2017supervised} and clustering~\cite{keraghel2024beyond,petukhova2024text}, 
which measures the similarity of semantic details between texts, emphasizing homogeneous symmetric relationships. 
STS and NLI are particularly useful in scenarios such as detecting duplicate content, recognizing paraphrases, query rewriting, or logic reasoning. 
Similarly, text clustering aims to group semantically close documents, which is critical for discovering underlying themes, organizing large text collections, and more. 
These cases collectively demonstrate that semantic similarity is a crucial aspect of text embeddings.

\subsubsection{Semantic Relevance (SR)}\label{sec:sr}
This type corresponds to IR~\cite{manning2009introduction,guo2022semantic} and QA~\cite{DBLP:journals/tacl/KwiatkowskiPRCP19,karpukhin-etal-2020-dense} mainly, 
where the goal is to search relevant documents or passages given a query, for instance, finding a text that can answer a specific question.
IR has been one of the most widely applied topics for text embeddings. 
This task is typically formulated as a scoring problem, where documents with the highest semantic relevance scores are returned as candidates for the query.
Retrieval can be employed in various scenarios, such as question answering~\cite{karpukhin-etal-2020-dense}, prompt retrieval \cite{cheng-etal-2023-uprise}, RAG~\cite{gao2023retrieval,DBLP:conf/naacl/ZhaoZSHLLHZ25,DBLP:conf/nips/LewisPPPKGKLYR020}, LLM memory~\cite{park2023generative,zhong2024memorybank}, fact verification~\cite{DBLP:conf/naacl/ThorneVCM18} \etc

\subsubsection{Semantic Encoding (SE)}\label{sec:se}
This line of applications involves using text embeddings as feature inputs for various downstream models, such as text classifiers~\cite{da2023text}, semantic reasoning systems~\cite{cheng2024xrag}, and image generators~\cite{saharia2022photorealistic}. 
Compared with traditional features, text embeddings are more expressive with reduced dimensions and high-level textual semantics, leading to better performance in downstream tasks~\cite{da2023text}. 
The RAG models could also benefit from text embeddings, where the retrieved documents could be replaced with tailored embeddings as inputs for LLMs~\cite{rau2024context,cheng2024xrag}.
\subsubsection{Hybrid Combination}\label{sec:hc}
There are special cases where different types of applications are combined~\cite{cheng2024xrag,DBLP:conf/acl/LiHLZYYZS25,cui2025think}.
For example, xRAG~\cite{cheng2024xrag} suggests directly inputting the retrieved text embeddings into the LLMs, where the embeddings are used both for text retrieval and as feature inputs.
As the capabilities of GPTE continue to evolve, we expect to see more hybrid applications in the future, largely due to the increasingly complex real-world scenarios that we are already able to resolve.
\begin{wrapfigure}{R}{0.375\linewidth}
    \centering
    \includegraphics[width=.99\linewidth]{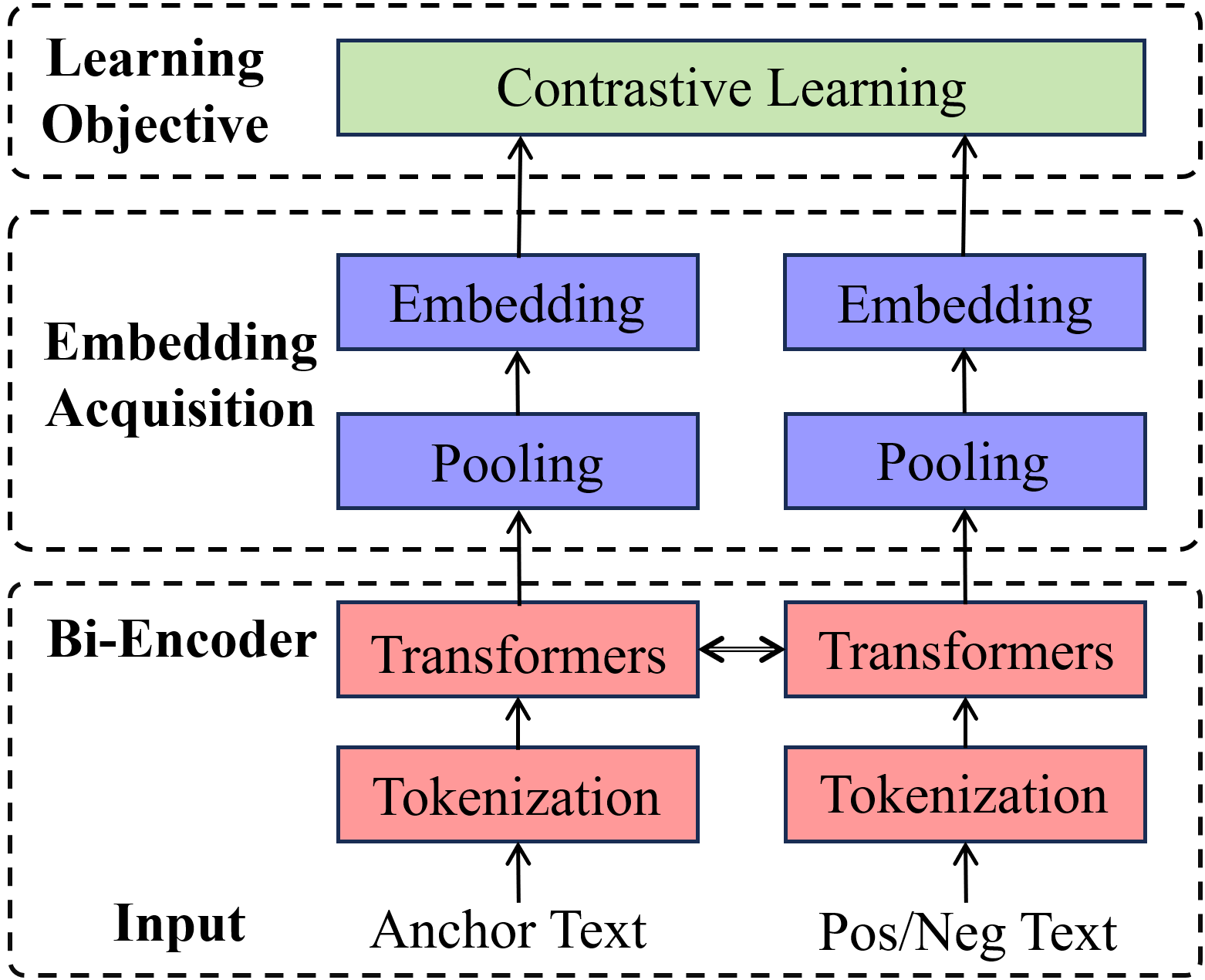}
    \caption{The typical architecture and training manner of GPTE models.}
    \label{fig:workflow}
\end{wrapfigure}
\subsection{The General Architecture of GPTE}\label{sec:common:architecture}
The task of text embedding requires a sophisticated model to possess a deep understanding of text.
A straightforward approach involves constructing a neural network from scratch to encode the input sequence of words into a fixed-size embedding.
The central challenge then lies in effectively learning the model parameters.
Inspired by the success of general-purpose PLM backbones in NLP, GPTE has garnered considerable attention.
These embeddings offer strong generalization for real-world, open-domain settings and can be further fine-tuned for specific tasks, aligning with the pretrain-then-finetune paradigm of PLMs. 
Consequently, parameter learning typically involves multiple stages, where we focus on GPTE, ignoring the last step of SFT for specific tasks.
Figure~\ref{fig:workflow} illustrates the mainstream architecture for pretraining GPTE through a supervised approach.
Typically, the textual word sequence is fed into a well-established PLM backbone (the key neural network is transformers), generating hidden contextual representations of words. 
Following this, a pooling step aggregates these word-level hidden vectors into a single vector, yielding the embedding form of the input text. 
Once the embedding network is ready, the subsequent phase involves further optimization beyond the PLM's initial capabilities, for which contrastive learning (CL) is the widely-accepted supervision objective~\cite{DBLP:conf/coling/ChenZS022,DBLP:conf/naacl/NishikawaRYTE22,DBLP:conf/acl/KleinN22,DBLP:conf/coling/WuGSHWH22,DBLP:conf/acl/ZhangZWXLZ22}.
This learning process can be self-supervised, weakly-supervised, or high-quality supervised through bi-encoder semantic computation.
Despite potential innovations such as reinforcement learning that could be exploited, this supervised paradigm remains the predominant architecture for nearly all current models.

Formally, the CL aims to optimize GPTE through text-pair similarity calibration. 
Given an anchor text $x$ (often the query text in IR), first, we provide one positive text $y^+$ which is relevant to $x$.
Based on the positive pair, then we construct a set of negative (e.g., irrelevant) samples $y^{-}_{i}$ ($i \in [1, N]$) of $x$. 
Finally,  the standard InfoNCE loss~\cite{oord2018representation} is adopted, maximizing the agreement of positive pairs and meanwhile minimizing that of negative pairs:
\begin{equation}
\mathcal{L}_\text{InfoNCE} = - \log \frac{ e^{s(\bm{x}, \bm{y}^+) / \tau} }{
    e^{s(\bm{x}, \bm{y}^+) / \tau} +\sum_{i=1}^{N}e^{s(\bm{x}, \bm{y}_i^{-})/ \tau}},
\end{equation}
where $\bm{x}$ and $\bm{y}$ denote the embeddings of $x$ and $y$ calculated by the embedding network, 
$s(\bm{x}, \bm{y})$ is the similarity function, normally cosine similarity or dot product, and $\tau$ is the temperature.
By design, CL encourages the embedding space to bring anchor-positive pairs closer while pushing anchor-negative pairs farther apart.
Besides the above standard CL, there are also several strategies to better CL optimization, e.g., margin-based CL~\cite{DBLP:conf/acl/ZhangZWXLZ22} and momentum smoothing~\cite{DBLP:conf/coling/0002GZHWH22,DBLP:conf/icse/ShiWGDZHZS23}.  
For these studies, we will not discuss them as they are not the main points in this survey.

\begin{table*}[t]
\centering
\footnotesize
\tabcolsep=0.1cm
\def\arraystretch{1.55}
\caption{Text embedding training datasets.}
\begin{tabularx}{1.00\textwidth}{c|c|X}
\hline
\textbf{Quality} & \textbf{Type} & \multicolumn{1}{c}{\bf Datasets} \\
\hline

\multirow{5}{*}{Low} 

  &  \multirow{3}{*}{SS}  & Amazon Reviews~\cite{DBLP:journals/corr/abs-2403-03952}, WikiHow~\cite{DBLP:journals/corr/abs-1810-09305}, NLLB~\cite{team2022language,DBLP:conf/emnlp/HeffernanCS22}, MNLI~\cite{DBLP:conf/naacl/WilliamsNB18}, SNLI~\cite{bowman2015large}, \href{https://huggingface.co/datasets/mteb/raw_arxiv}{ArXiv-Clustering}, \href{https://huggingface.co/datasets/mteb/raw_biorxiv}{Biorxiv-Clustering}, \href{https://huggingface.co/datasets/mteb/raw_medrxiv}{Medrxiv-Clustering}, Reddit-Clustering~\cite{DBLP:journals/corr/abs-2104-07081,sentence-transformers-reddit}, StackExchange-Clustering~\cite{DBLP:journals/corr/abs-2104-07081,sentence-transformers-stackexchange}, XL-Sum~\cite{DBLP:conf/acl/HasanBIMLKRS21} \\
  \cline{2-3}
  
  & \multirow{3}{*}{SR} & MS MARCO~\cite{DBLP:conf/nips/NguyenRSGTMD16}, \href{https://commoncrawl.org/}{Common Crawl}, ClueWeb~\cite{overwijk2022clueweb2210billionweb}, GooAQ~\cite{DBLP:conf/emnlp/KhashabiNKSHC21}, \href{https://huggingface.co/datasets/sentence-transformers/embedding-training-data}{Yahoo Answers}, Wikipedia~\cite{wikidump}, \href{https://huggingface.co/datasets/teven/stackexchange}{Stack Exchange}, PAQ~\cite{DBLP:journals/tacl/LewisWLMKPSR21}, CC-News~\cite{DBLP:conf/isiwi/HamborgMBG17}, \href{https://huggingface.co/datasets/sentence-transformers/reddit}{Reddit}, \href{https://huggingface.co/datasets/sentence-transformers/s2orc}{S2ORC}, x3P~\cite{DBLP:conf/acl/MuennighoffWSRB23}, pubMedQA~\cite{DBLP:conf/emnlp/JinDLCL19}, \href{https://huggingface.co/datasets/TitanMLData/arxiv_qa}{arxiv\_qa}, SQuAD~\cite{DBLP:conf/emnlp/RajpurkarZLL16}, DBPedia~\cite{thakur2beir}. BERRI~\cite{DBLP:conf/acl/AsaiSL0I0HY23}, MEDI~\cite{su-etal-2023-one} \\
  \cline{2-3}

\hline
\multirow{6}{*}{High} 
  & \multirow{2}{*}{SS} &  LCSTS~\cite{DBLP:conf/emnlp/HuCZ15}, STS12~\cite{DBLP:journals/sigir/BalogCVHMRSST12}, STS22~\cite{DBLP:conf/semeval/ChenZCFGGHJS22}, STSB~\cite{DBLP:conf/semeval/CerDALS17}, ContractNLI~\cite{DBLP:conf/emnlp/KoreedaM21}, MultiNLI~\cite{DBLP:conf/naacl/WilliamsNB18}, Quora~\cite{quora-question-pairs}, WikiAnswers~\cite{DBLP:conf/kdd/FaderZE14}, SimCSE NLI~\cite{DBLP:conf/emnlp/GaoYC21}, \href{https://huggingface.co/datasets/C-MTEB/AFQMC}{AFQMC}, \href{https://huggingface.co/datasets/C-MTEB/ATEC}{ATEC}, \href{https://huggingface.co/datasets/C-MTEB/BQ}{BQ}, CAIL2029-SCM~\cite{DBLP:journals/corr/abs-1911-08962}, \href{https://www.luge.ai/\#/luge/dataDetail?id=39}{CINLID}, ChineseSTS~\cite{ChineseSTS}
  \\
  \cline{2-3}
  
  & \multirow{3}{*}{SR} & FEVER~\cite{DBLP:conf/naacl/ThorneVCM18}, HotpotQA~\cite{DBLP:conf/emnlp/Yang0ZBCSM18}, Natural Questions~\cite{DBLP:journals/tacl/KwiatkowskiPRCP19},  WebQA~\cite{DBLP:conf/cvpr/ChangCNGSB22}, SciFact~\cite{DBLP:conf/emnlp/WaddenLLWZCH20}, ChatMed-Dataset~\cite{ChatMed}, LIMA~\cite{DBLP:conf/nips/ZhouLX0SMMEYYZG23}, T2Ranking~\cite{DBLP:conf/sigir/XieDWLYG0LL0M23}, RefGPT~\cite{DBLP:conf/emnlp/YangYFYWWZ23}, TriviaQA~\cite{DBLP:conf/acl/JoshiCWZ17}, FiQA~\cite{DBLP:conf/www/MaiaHFDMZB18}, BioASQ~\cite{DBLP:journals/bmcbi/TsatsaronisBMPZ15}, DuReader~\cite{DBLP:conf/acl/HeLLLZXLWWSLWW18}, DuReader$_\mathrm{checklist}$~\cite{DBLP:conf/acl/TangL0H0020}
  \\
  \cline{2-3}
    
  & \multirow{3}{*}{SE}  & EmotionClassification~\cite{DBLP:conf/emnlp/SaraviaLHWC18},  \href{https://huggingface.co/datasets/mteb/toxic_conversations_50k}{ToxicConversations}, \href{https://huggingface.co/datasets/mteb/tweet_sentiment_extraction}{TweetSentimentExtraction}, AmazonPolarity~\cite{DBLP:conf/recsys/McAuleyL13}, IMDB~\cite{DBLP:conf/acl/MaasDPHNP11}, banking77~\cite{DBLP:journals/corr/abs-2003-04807}, CSL~\cite{DBLP:conf/coling/LiZ0S0MZ22}, THUCNews~\cite{thuctc}, \href{https://huggingface.co/datasets/fenffef/tnews}{TNews}, \href{https://huggingface.co/datasets/C-MTEB/JDReview-classification}{JDReview}, IFlyTek~\cite{DBLP:journals/corr/abs-2202-10974}, \href{https://huggingface.co/datasets/C-MTEB/OnlineShopping-classification}{OnlineShopping}, \href{https://huggingface.co/datasets/C-MTEB/waimai-classification}{Waimai}  \\
\hline
\end{tabularx}
\label{tab:text_emb_training_data}
\end{table*}

\subsection{Training Dataset}\label{sec:training:data}
Previous studies often adopt multi-task supervision to train text embeddings using diverse task-specific 
objectives—including classification~\cite{kim-2014-convolutional,howard2018universal,da2023text}, STS~\cite{DBLP:conf/emnlp/ReimersG19,hill2016learning}, question answering~\cite{shi2024replug,guu2020retrieval}, and text retrieval~\cite{karpukhin-etal-2020-dense,li2023towards,li2024making}—spanning 
various GPTE-based applications with supervised datasets available.  
In fact, nearly all of these datasets can be reformulated into text-pair formats, enabling CL to emerge as a universal paradigm for training GPTE as mentioned in Section \ref{sec:common:architecture}.  
First, Semantic similarity and relevance datasets such as the manually-refined MS MARCO~\cite{DBLP:conf/nips/NguyenRSGTMD16}, PQA~\cite{DBLP:journals/tacl/LewisWLMKPSR21}, MNLI~\cite{nangia2017repeval}, SNLI~\cite{bowman2015large}, DBPedia~\cite{thakur2beir}, are naturally well-suited for CL pretraining.
Furthermore, the datasets targeting feature representation tasks can also be adapted for CL. 
For instance, in text classification, texts sharing the same label can form positive pairs, and in semantic reasoning tasks, instruction–response pairs can serve the same purpose. 
In addition to these high-quality resources, relatively lower-quality yet large-scale corpora crawled from the web are also widely used. 
These include title-body pairs from regular web pages, title-abstract pairs from academic papers, post–comment pairs from social media, question–answer pairs from community QA forums, text-code pairs from GitHub, and other text pairs generated via retrieval. 
Table \ref{tab:text_emb_training_data} presents a summary of these representative training datasets.
\subsection{GPTE Evaluation}\label{sec:gpte:eval}
The evaluation of text embeddings is critically important, as it directly guides the development of GPTE. 
In the early stage, text embeddings were evaluated independently using various benchmark datasets. 
One of the earliest datasets, SentEval~\cite{DBLP:conf/lrec/ConneauK18}, was introduced to evaluate embeddings on tasks such as classification, STS, and NLI. 
Subsequently, USEB~\cite{DBLP:journals/corr/abs-2104-06979} was proposed to focus on the reranking task, and BEIR~\cite{thakur2beir} was constructed for zero-shot information retrieval evaluation. 
Further, SciRepEval~\cite{DBLP:conf/emnlp/SinghDCDF23} was developed for evaluating scientific document representations, covering 24 realistic tasks across classification, regression, ranking, and search.
Recently, MTEB has emerged as a unified benchmark encompassing all major types of embedding tasks, serving as the primary framework for evaluating various GPTE models. 
To date, MTEB has undergone several iterations and includes a wide array of datasets across different languages, e.g., 56 datasets for the English language~\cite{muennighoff-etal-2023-mteb}, 35 datasets for Chinese~\cite{xiao2023c}, 25 datasets for French~\cite{ciancone2024mteb}, 17 datasets for Polish~\cite{DBLP:journals/corr/abs-2405-10138}, 12 code retrieval tasks for MTEB (code)~\cite{enevoldsen2025mmteb}, etc.
Building upon MTEB~\cite{muennighoff-etal-2023-mteb}, MMTEB~\cite{enevoldsen2025mmteb} extends multilingual evaluation to over 250 languages, offering a diverse set of more than 500 systematically curated tasks for evaluating text embedding models.
The construction of MTEB and MMTEB is still ongoing, continuously expanding to include broader task types, languages, and domains, with increasing levels of difficulty.
Notably, specific efforts for language expansion include C-MTEB~\cite{xiao2023c} for Chinese, SEB~\cite{DBLP:conf/nips/EnevoldsenKMN24} for Scandinavian languages, PIRB~\cite{DBLP:conf/coling/DadasPP24} for Polish, and ruMTEB~\cite{DBLP:conf/naacl/SnegirevTMFA25} for Russian, BEIR-NL~\cite{DBLP:journals/corr/abs-2412-08329} for Dutch, and Hindi-BEIR~\cite{DBLP:journals/corr/abs-2408-09437} for Hindi.

\begin{table*}
\centering
\Large
\caption{
Representative open-source GPTE models (English-centered),
where only models released on Huggingface with detailed and clear documentation are listed.
New Abbreviations: M.S. = multi‑stage training; MNTP = masked next token prediction training; COS = cosine objective.
}
\def\arraystretch{0.975}
\resizebox{\textwidth}{!}{
\tabcolsep=0.105cm
\begin{tabular}{cccccccccccc}
\toprule
 \textbf{Model} &  \textbf{PLM}  &  \textbf{Release} &  \textbf{Param.} &  \textbf{Embed.} &  \textbf{Max} &  \textbf{Embed.} & \multicolumn{2}{c}{Learning} \\ \cline{8-9}
 \textbf{Series}  &  \textbf{Base}  &  \textbf{Time} &  \textbf{Size} &  \textbf{Dim.} &  \textbf{Tokens} &  \textbf{Acqui.} & \multicolumn{1}{c|}{\bf M.S.} &  \textbf{Object.} \\
\midrule
\multirow{5}{*}{\shortstack[c]{ GTE\\ (Advancing)}  } & BERT & 2023-07 \cite{li2023towards} & 33M$\sim$335M & 384$\sim$1024 & 512 & mean & $\checkmark$ & CL   \\
   & GTE-MLM & 2024-04 \cite{DBLP:conf/emnlp/ZhangZLXDTLYXHZ24} & 137M/434M & 768/1024 & 8192 & first & $\checkmark$ & CL+MRL+LEX  \\
   & Qwen1.5 & 2024-04 \cite{li2023towards} & 7.72B & 4096 & 32768 & last & $\checkmark$ & CL+MRL+LEX  \\ 
   & Qwen2 & 2024-06 \cite{li2023towards} & 1.78B/7.61B & 1536/3584 & 32768 & last & $\checkmark$ & CL+MRL+LEX  \\
   & Qwen3 & 2025-06 \cite{qwen3embedding} & 0.6B$\sim$8B & 1024$\sim$4096 & 32768 & last & $\checkmark$ & CL+MRL+LEX  \\
   \hline
\multirow{ 4}{*}{E5 } & BERT & 2022-12 \cite{wang2022text} & 33M$\sim$335M & 384$\sim$1024 & 512 & mean & $\checkmark$  & CL+KD  \\
  & BERT & 2023-05 \cite{wang2022text} & 33M$\sim$335M & 384$\sim$1024 & 512 & mean & $\checkmark$ & CL+KD  \\
  & BERT  & 2024-04 \cite{zhu2024longembed} & 109M & 768 & 4096 & mean & $\checkmark$ & CL \\ 
  & Mistral  & 2023-12 \cite{wang2023improving} & 7.11B & 4096 & 4096 &  last & $\times$ & CL \\
  \hline
\multirow{ 4}{*}{BGE } & BERT & 2023-08 \cite{xiao2023c} & 33M$\sim$335M & 384$\sim$1024 & 512 & first & $\checkmark$  & CL   \\
  & BERT & 2023-09 \cite{xiao2023c} & 33M$\sim$335M & 384$\sim$1024 & 512 & first & $\checkmark$ & CL  \\
  & Mistral & 2024-07 \cite{li2024making} & 7.11B & 4096 & 4096  & last & $\checkmark$  & CL+KD \\
  & Gemma  & 2024-07 \cite{li2024making} & 9.24B & 3584 & 8192 & last & $\checkmark$ & CL+KD \\
  \hline 
\multirow{3}{*}{Jina } & T5-Enc  & 2023-07 \cite{gunther2023jina} & 14M$\sim$335M & 312$\sim$1024 & 512 & mean & $\checkmark$ & CL \\
  & JinaBERT & 2023-10 \cite{gunther2023jina2} & 33M/137M & 512/768 & 8192  & mean & $\checkmark$ & CL \\
  & XLM-R & 2024-09  \cite{sturua2024jina3} & 572M & 1024 & 8192 & mean &  $\checkmark$ & CL+MRL \\
  \hline
\multirow{3}{*}{Nomic } & NomicBERT & 2024-02 \cite{nussbaum2024nomic} & 137M & 768 & 8192 & first & $\checkmark$ & CL  \\
   & NomicBERT  & 2024-02 \cite{nussbaum2024nomic} & 137M & 768 & 8192 & first & $\checkmark$  & CL+MRL  \\
   & XLM-R MoE & 2025-02 \cite{nussbaum2025nomicmoe} & 305M & 768 & 512 & first & $\checkmark$  & CL+MRL \\   
  \hline
\multirow{3}{*}{LLM2Vec} & LLaMA-2 & 2024-04 \cite{behnamghader2024llm2vec} & 7B & 4096 & 4096 & mean & $\checkmark$ & CL+MNTP \\
 & LLaMA-3 & 2024-04 \cite{behnamghader2024llm2vec} & 8B & 4096 & 8192 & mean & $\checkmark$ & CL+MNTP \\ 
 & Mistral & 2024-04 \cite{behnamghader2024llm2vec} & 7B & 4096 & 32768 & mean & $\checkmark$ & CL+MNTP \\ \hline 
\multirow{2}{*}{KaLM} & Qwen2 & 2024-12 \cite{hu2025kalm} & 0.49B & 1024 & 512 & last & $\times$ & CL+MRL \\
& Qwen2 & 2025-06 \cite{zhao2025kalmembeddingv2} & 0.49B & 1024 & 512 & last & $\checkmark$ & CL+MRL \\ \hline
 \multirow{2}{*}{ST5}   & T5-Enc  & 2021-12 \cite{ni-etal-2022-sentence} & 110M$\sim$4.8B & 768/1024 & 512 & mean & $\checkmark$ & CL \\
        & T5-EncDec  & 2021-12 \cite{ni-etal-2022-sentence} & 220M$\sim$11B & 768/1024 & 512 & Dec-first & $\checkmark$ & CL \\
   \hline
   & BERT & 2024-04 \cite{merrick2024arctic} & 22M$\sim$335M & 384$\sim$1024 & 512 & first & $\checkmark$ & CL  \\
  snowflake & NomicBERT & 2024-04 \cite{merrick2024arctic} & 137M & 768 & 8192 & first & $\checkmark$ & CL \\
arctic & GTE-MLM & 2024-12 \cite{yu2024arctic2} & 304M & 768 & 8192 & first &  $\checkmark$ & CL+MRL \\
& XLM-R  & 2024-12 \cite{yu2024arctic2} & 568M & 1024 & 8192 & first &  $\checkmark$ & CL+MRL \\
  \hline 
\multirow{2}{*}{CDE} & NomicBERT & 2024-10 \cite{morris2024cde} & 281M & 768 & 512 & mean & $\checkmark$ & CL \\ 
 & ModernBERT & 2025-01 \cite{morris2024cde} & 306M & 768 & 512 & mean & $\checkmark$ & CL \\ 
  \hline  
\multirow{2}{*}{GritLM} & Mistral & 2024-02 \cite{muennighoff2024generative} & 7.24B & 4096 & 4096  & mean &  $\times$ & CL+NTP \\
    & Mistral & 2024-02 \cite{muennighoff2024generative} & 46.7B (8x7B) & 4096 & 32768 & mean & $\times$  & CL+NTP\\
  \hline       
\multirow{2}{*}{NV-Embed} & Mistral & 2024-05 \cite{lee2024nv} & 7.85B & 4096 & 4096 & Latent & $\checkmark$ & CL \\
  & Mistral & 2024-08 \cite{lee2024nv} & 7.85B & 4096 & 32768 & Latent & $\checkmark$  & CL \\
  \hline  
\multirow{2}{*}{DiffCSE} &   BERT & 2022-05 \cite{DBLP:conf/naacl/ChuangDLZCS0YKG22} & 110M & 768 & 512 & first & $\checkmark$ & CL+RTD  \\
     &   RoBERTa & 2022-05 \cite{DBLP:conf/naacl/ChuangDLZCS0YKG22} & 125M & 768 & 512 & first & $\checkmark$ & CL+RTD  \\
  \hline 
\multirow{2}{*}{EASE}   &   BERT & 2022-12 \cite{DBLP:conf/naacl/NishikawaRYTE22} & 110M & 768 & 512 & first & $\checkmark$ & CL  \\
     &   RoBERTa & 2022-12 \cite{DBLP:conf/naacl/NishikawaRYTE22} & 125M & 768 & 512 & first & $\checkmark$ & CL  \\
  \hline   
DeCLUTR & BERT & 2022-08 \cite{DBLP:conf/acl/GiorgiNWB20} & 125M & 768 & 512 & first & $\checkmark$ & CL+MLM  \\     
UAE & BERT & 2023-12 \cite{li2023angle} & 335M & 1024 & 512 & first & $\checkmark$ & CL+COS+AnglE  \\
Granite & RoBERTa & 2024-12 \cite{awasthy2025granite} & 30M$\sim$278M & 384$\sim$768 & 512 & first & $\checkmark$ & CL \\ 
 GTR   & T5-Enc  & 2021-12 \cite{DBLP:conf/emnlp/Ni0LDAMZLHCY22} & 110M$\sim$4.8B & 768/1024 & 512 & mean & $\checkmark$ & CL \\
Instructor  & GTR(T5-Enc)  & 2022-12 \cite{su-etal-2023-one} & 110M/335M & 768/1024 & 512 & mean & $\checkmark$ & CL \\
SGPT & Bloom & 2022-08 \cite{muennighoff2022sgpt} & 7.07B & 4096 & 512 &  mean & $\times$ & CL \\
Udever & Bloom & 2023-10 \cite{zhang2023language} & 560M$\sim$7.07B & 1024$\sim$4096 & 512 & last & $\times$  & CL \\
SFR-Embed & Mistral & 2024-01 \cite{SFRAIResearch2024} & 7.11B & 4096 & 32768 & last & $\checkmark$ & CL \\
Echo & Mistral & 2024-02 \cite{springer2024repetition} & 7.11B & 4096 & 32768 & last & $\times$ & CL \\
Linq-Embed & Mistral & 2024-05 \cite{choi2024linq} & 7.11B & 4096 & 32768 & last & $\checkmark$ & CL \\
SPEED & Mistral & 2024-11 \cite{chen2024little} & 7.11B & 4096 & 32768 & last & $\times$  & CL \\
\bottomrule
\end{tabular}
}

\label{tab:embed-models}
\end{table*}

\section{The Basic Roles of PLMs}
\label{sec:role}
Since the introduction of ELMo \cite{DBLP:conf/naacl/PetersNIGCLZ18}, PLMs have remained a central focus in the NLP community due to their remarkable performance.
Models such as BERT \cite{devlin-etal-2019-bert} and GPT \cite{radford2018improving} have significantly advanced the overall capabilities of NLP \cite{WANG202351}. 
In the context of text embeddings, PLMs, particularly BERT-like architectures, have become the primary approach for enhancing embedding quality \cite{DBLP:conf/emnlp/ReimersG19,karpukhin-etal-2020-dense}.
Given that PLMs are inherently general-purpose and follow the pretrain-then-finetune paradigm, text embeddings derived from them naturally inherit this generalizability \cite{wang2022text,su-etal-2023-one}.
Here, we build upon the aforementioned general architecture in Section \ref{sec:background} to examine the role of PLMs in advancing GPTE.
Table \ref{tab:embed-models} shows the representative GPTE models with their detailed architecture information.

This section is organized as follows. 
First, we provide a detailed overview of how text embeddings are extracted from PLMs (\S\ref{sec:embedding:acquire}), which serves as the foundational step in leveraging PLMs for GPTE.
Next, we examine approaches for improving embedding expressibility in the context of different PLMs for GPTE (\S\ref{sec:role:express}).
Third, we present recent advancements in optimization techniques, describing various learning objectives to obtain better GPTE (\S\ref{sec:role:opt}). 
Fourthly,  we describe the development of data synthesis for the training and evaluation with the support of PLMs (\S\ref{sec:role:data}).
Finally, we compare GPTE across various types of PLMs, highlighting the key factors that influence their performance (\S\ref{sec:role:choice}).

\subsection{Text Embedding Acquisition}
\label{sec:embedding:acquire}
To date, most PLMs adhere to a unified transformer-based architecture in which word-level representations are interconnected through attention mechanisms, allowing each token to attend to others within the sequence and thereby generating contextualized word representations~\cite{DBLP:conf/nips/VaswaniSPUJGKP17}.
Encoder-based PLMs employ bidirectional attention to aggregate contextual information~\cite{devlin-etal-2019-bert}, 
while decoder-based PLMs utilize causal (unidirectional) attention to construct representations in an autoregressive manner~\cite{radford2018improving}. 
Encoder-decoder PLMs use a mixture of them.  
Regardless of the underlying architecture, a pooling strategy is typically employed to produce a fixed-size vector that serves as the embedding of the entire text:
$\bm{x} = \sum_{i=1}^{n} \bm{\alpha}_i \bm{x}_i$,
where $\bm{x}_i$ is the hidden vector representation of word $w_i$ in text $x$.
The hidden vector is often the last layer neural representation of PLMs. 
When using encoder-based PLMs, e.g.,  BERT~\cite{devlin-etal-2019-bert} and RoBERTa~\cite{liu2019roberta}, 
the representation of the first token (i.e., $\langle \text{CLS} \rangle$, where $\alpha_1 = 1$ and all other $\alpha_i = 0$) is typically adopted as the full text representation.
These models leverage bidirectional attention, allowing each token to attend to all others in the input sequence 
and thereby enabling the first-token representation to capture rich contextual information from the entire text.
However, relying solely on the $\langle \text{CLS} \rangle$ token may introduce bias. 
To address this, mean pooling (i.e., $\alpha_i = \frac{1}{n}$) is often employed as a more balanced alternative. 
In some scenarios, max pooling is also used, and in particular, attentive pooling, e.g., incorporating an additional latent attention layer for representation aggregation, can further enhance the quality of the resulting text embedding~\cite{lee2024nv}.
For encoder-decoder models such as T5~\cite{raffel2020exploring}, pooling is typically performed in a manner similar to encoder-based PLMs.
By feeding the full text into the encoder component only, the same pooling strategies, such as using the $\langle \text{CLS} \rangle$ token or mean pooling, can be employed to obtain text embeddings.
Additionally, there are several studies that attempt to leverage the decoder component.
By using the start token in the decoder as the anchor for embedding extraction while ignoring the remainder of the decoding process, it is feasible to produce reasonable full-text embeddings~\cite{ni-etal-2022-sentence}.
Decoder-only LLMs have revolutionized the field of NLP, achieving remarkable success across a wide range of tasks such as machine translation, question answering, and semantic reasoning.
However, their inherent causal attention mechanism presents a unique challenge for text embedding, which makes it challenging to derive a single comprehensive vector that fully captures the semantic meaning of the input text.
Only the final token possesses the complete contextual information, making it the most reasonable choice for representation.
As a result, last-token pooling (i.e., $\alpha_n = 1$, with all other weights being zero) is the most common strategy for extracting text embeddings from decoder-only LLM models.
However, relying solely on the final token’s representation may fail to capture the exhaustive semantic nuances of longer and more complex texts.
To overcome this limitation, researchers have begun exploring more sophisticated aggregation strategies.
One natural extension involves averaging the representations of multiple tokens from the tail of the sequence~\cite{muennighoff2022sgpt}, 
i.e.,   $\bm{x} = \sum_{i=k}^{n} \frac{1}{n-k+1}\bm{x}_i$, 	where $k$ marks the starting index of the tail segment used for aggregation. 
This ``partial pooling'' approach seeks to incorporate a broader contextual window, potentially yielding more robust and semantically expressive text embeddings.
Several studies argue that relying solely on the final-layer token representations in PLMs may be insufficient for obtaining robust text embeddings~\cite{su2021whitening}.
To address this, a stack of multiple layers (i.e., Top-K layers) can be leveraged to capture richer contextual information. 
A straightforward approach to combine these layers is simple average aggregation, while more advanced methods involve trainable multi-layer pooling modules to integrate information across layers~\cite{oh-etal-2022-dont}.
After this multi-layer aggregation, another token(word)-wise pooling is applied over the resulting word representations to generate the ultimate text embeddings.

\subsection{Expressibility Improvement}
\label{sec:role:express}
\subsubsection{Long-Context Modeling}\label{sec:long:modeling}
Early BERT-style encoder models typically support sequences of up to 512 tokens, which falls short given the demand for longer text processing.
To overcome this limitation, many studies~\cite{neelakantan2022text,wang2023improving,muennighoff2024generative} have adopted PLM backbones that natively support extended contexts such as recently-released decoder-only LLMs. 
Despite decoder-only models having gained considerable traction due to their inherent capacity for long-context understanding, substantial research continues to focus on extending the context length of existing encoder-based models~\cite{gunther2023jina,nussbaum2024nomic,DBLP:conf/coling/YanoFFTW24,zhu2024longembed,chen2024bge},
since these GPTE models still remain highly practical and effective in real-world applications.
These efforts generally follow two main strategies: plug-and-play augmentation or full-scale pre-training from scratch.
{Plug-and-Play Augmentation.}
One quick solution for extending context length is to make low-cost adaptations to existing models. 
For instance, BGE-M3~\cite{chen2024bge} significantly extends XLM-RoBERTa's context window to 8192 tokens.
It achieves this by first replicating the learned position embeddings multiple times, and then continues pre-training on longer sequences using RetroMAE~\cite{xiao-etal-2022-retromae}.
Alternatively, LongEmbed~\cite{zhu2024longembed} and Jina-Embeddings-v3~\cite{sturua2024jina3} demonstrate how BERT and RoBERTa can be enhanced for longer contexts by integrating RoPE~\cite{su2024roformer} and performing contrastive pre-training.
These approaches demonstrate that even without architectural modifications, existing encoders can be efficiently adapted to support much longer contexts.

{Pre-Training from Scratch.}
Beyond adapting existing PLMs, many efforts have focused on modifying the transformer architecture to develop new encoder-only models from scratch that support longer contexts.
These approaches typically build on encoder-based PLMs like BERT due to their relatively low training costs. 
For instance, MosaicBERT~\cite{portes2024mosaicbert} and JinaBERT~\cite{gunther2023jina2} replace the positional embedding of BERT with Alibi positional bias~\cite{press2021train}, enabling robust inference on longer sequences than those seen during training\footnote{MosaicBERT~\cite{portes2024mosaicbert} explored these improvements first, but they did not develop embedding models.};
In contrast, nomicBERT~\cite{nussbaum2024nomic}, GTE-MLM~\cite{DBLP:conf/coling/YanoFFTW24} and ModernBERT~\cite{warner2024smarter} adopt Rotary Position Embeddings (RoPE)~\cite{su2024roformer}, 
which have generally shown superior performance over Alibi.
Additionally, some research has also explored non-transformer-based models to support long contexts, like M2-BERT~\cite{saad2024benchmarking}.

\subsubsection{Prompt-Informed Embedding}\label{sec:prompt:emb}
Most methods of text embedding acquisition discussed in Section~\ref{sec:embedding:acquire} essentially perform a selective averaging of token-wise embeddings, effectively producing an extractive summarization of the full text.
This perspective introduces a novel paradigm for GPTE based on text summarization or compression.
The core idea is to condense the original text into a few representative words and then use the embeddings of these words to represent the full text. 
This paradigm can be effectively realized through prompt learning, a technique widely used in PLM reasoning.
PromptBERT~\cite{jiang-etal-2022-promptbert} is among the earliest attempts to leverage prompts for inducing text embeddings.  
With a template such as ``[TEXT] means [MASK] .'', where ``[TEXT]'' denotes the input text and ``[MASK]'' is the summarized special symbol, we can use ``[MASK]'' to extract text embeddings.
In addition, soft prompts are also investigated in the work of PromCSE~\cite{DBLP:conf/emnlp/JiangZW22}, while unlike PromptBERT, this work aims to learn better GPTE embeddings from the optimization perspective.   
The method has received great attention in the era of auto-regressive LLMs~\cite{cheng-etal-2023-uprise,DBLP:conf/iclr/DaiZMLNLBGHC23,zhuang-etal-2024-promptreps,DBLP:conf/iclr/WellerDLPZH25}.
PCoTEOL~\cite{zhang2024simple} uses a more sophisticated prompt with chain-of-thought: ``After thinking step by step, this sentence : [TEXT] means in one word: [MASK].''
Echo~\cite{springer2024repetition} proposes a repetition-style prompt: ``Rewrite the sentence: [TEXT], rewritten sentence: [TEXT]'', where both ``[TEXT]'' placeholders contain the same text.
The text embedding is obtained by averaging the token representations of the second occurrence.
In fact, the prompt can be highly flexible within this paradigm and can be naturally extended to instruction-following text embeddings, such as E5-mistral~\cite{wang2023improving}. 
GritLM~\cite{muennighoff2024generative} and GEM~\cite{zhang2025gem} further advance this line of work by adopting generalized prompt formats to unify the prompt-based embedding acquisition. 
``[prefix] [special tokens] [suffix]'' is an effective example, where ``[prefix]'' includes the input text along with an instruction, ``[special tokens]'' are a manually-defined symbol sequence used to extract the embeddings,
and ``[suffix]'' provides the corresponding response to the instruction.
Under this paradigm, we can exploit variable instruction-following tasks such as text reconstruction and summarization to obtain text embeddings.
Notably, even when only the ``[prefix]'' (i.e., the input text) and ``[special tokens]'' are retained, the resulting embeddings remain strong. 
This approach essentially treats embedding as a form of text compression, using a small set of ``[special tokens]'' to denote the semantics of the entire text~\cite{mu2023learning}.

\subsection{Parameter Optimization}
\label{sec:role:opt}
\subsubsection{Multi-Stage Training}\label{sec:multi:stage}
Directly using the PLM parameters is the most initial way for GPTE models with PLM backbones.
However, it is evident that the GPTE performance can be significantly enhanced through further parameter tuning.
Before, we have introduced that supervised contrastive learning with high-quality data is a common way for parameter tuning.
As previously discussed in Section~\ref{sec:background}, supervised CL on high-quality datasets remains a common and effective strategy for this purpose. 
With the rising popularity of weakly supervised and self-supervised paradigms, researchers have also begun to explore these approaches for GPTE, adopting multi-stage training pipelines to refine the text embedding quality progressively~\cite{wang2022text,li2023towards}.
The core idea of multi-stage training is to train the model step-by-step, from coarse to fine, taking full advantage of different types of data and gradually improving the model's generalization ability and task-specific performance.
Generally, we divide the data into two types: (1) Weakly-supervised data: with huge size (\eg billions of pairs) and relatively low costs, these data provide extensive semantic associations and contextual coverage, which helps models learn basic, generalized semantic representations. 
However, its annotation signal is noisy, and the task definition is vague.
(2) High-quality supervised data: with thousands to millions of pairs and high cost, these data deliver precise, high-quality task-specific signals that can directly guide the model in learning how to generate optimal embeddings under a particular task or instruction.
\textbf{The first pre-finetuning stage} begins by selecting a robust pre-trained model as the starting point and further enhancing its semantic representation abilities with massive weakly supervised datasets.
This process enables the model to learn broad correlations across diverse textual pairs, thereby establishing a strong basis for subsequent fine-tuning.
On top of the model after pre-finetuning, \textbf{the second finetuning stage} is performed using high-quality, task-specific, and instruction-specific fine-labeled data.
The core goal for the model is to generate different representations optimized for a specific task under different instructions, and distinguish between semantically highly similar but practically mismatched texts with fine-grained discrimination.

\subsubsection{Objectives Beyond Contrastive Learning}\label{sec:obj:beyong:cl}
Although the CL training has unified a number of GPTE tasks, such as classification, STS, NLI, retrieval, question-answering, instruction-following, etc., there are still other strategies to strengthen text embeddings.
For example, the original PLM-related objectives, such as masked language model (MLM), replaced token detection (RTD), next token prediction (NTP), and masked next‑token prediction (MNTP), 
can be adopted joint training with the GPTE objectives~\cite{DBLP:conf/emnlp/SchickS21a,DBLP:conf/naacl/ChuangDLZCS0YKG22,DBLP:conf/emnlp/WuGLHWH22,DBLP:conf/emnlp/WuZ22,muennighoff2024generative,behnamghader2024llm2vec}.
By the method, the GPTE can be greatly enhanced with better generalization capability.     
In addition, Matryoshka Representation Learning (MRL)~\cite{kusupati2022matryoshka} is one widely-adopted objective beyond CL, enabling the creation of hierarchical, multi-scale text embeddings.
The core idea is to encode coarse-grained information in the initial dimensions of the embedding vector and progressively add finer details in subsequent dimensions. 
Instead of optimizing a single loss value based on the full-sized embedding, MRL computes and sums the losses for various truncated versions of the embedding.
There are also several other cosine-based similarity objectives~\cite{su2022cosent} which differ from the CL objective.
The angle objective~\cite{li2023angle} (AnglE) is proposed to solve the saturation zone problem of cosine-based (COS) similarity objectives.
With a combination of the original objective of CL, it can optimize the angle difference in complex space to mitigate these adverse effects. 
Several studies introduce the lexicon-based sparse embeddings~\cite{formal2021splade,formal2022distillation} into the GPTE training, where each dimension of the vector corresponds to the weight of a token.
This lexicon-aligned, more interpretable sparse representation (LEX) could effectively enhance the retrieval performance for long text inputs~\cite{chen2024bge,zhuang-etal-2024-promptreps,DBLP:conf/emnlp/ZhangZLXDTLYXHZ24}.
Knowledge distillation (KD) from a cross-encoder (reranking) teacher model has also been explored~\cite{DBLP:conf/sigir/Cheng21,wang2022text,DBLP:conf/acl/SeonwooWSCLLXPO23,DBLP:conf/acl/LiuLWWWX0C023,chen2024bge}.
We could optimize the KL divergence loss towards soft labels from the teacher model to learn more precise semantic relations.
In addition to supervised approaches, there are several methods struggling to learn GPTE without relying on task-specific supervision. 
IS-BERT~\cite{DBLP:conf/emnlp/ZhangHLLB20} introduces a self-supervised objective based on mutual information maximization. 
BERT-flow~\cite{DBLP:conf/emnlp/LiZHWYL20} applies normalizing flows to transform the anisotropic embeddings from BERT into a smooth and isotropic Gaussian distribution. 
WhiteningBERT~\cite{DBLP:conf/emnlp/HuangTZLSGJD21} employs a whitening transformation, a simple yet effective linear transformation method, to improve embedding quality. 
SBERT-LP~\cite{DBLP:conf/emnlp/MinCY0L21} enhances semantic structure preservation by incorporating a locality-preserving loss. 
Meanwhile, SCD~\cite{DBLP:conf/acl/KleinN22} proposes a joint objective that combines self-contrastive learning with decorrelation, leveraging variations induced by different dropout patterns. 
These methods collectively highlight the potential of unsupervised and self-supervised strategies in improving the quality of text embeddings.
\subsubsection{Batch Learning}\label{sec:batch:learn}
Under the batch learning setting, training GPTE models with the CL objective batch differs from standard deep learning tasks.
The key reason lies in the construction of in-batch negatives for CL.
Given a batch of positive text pairs (i.e., $\langle q_1, d_1 \rangle, \langle q_2, d_2 \rangle, \cdots, \langle q_n, d_n \rangle$), we can randomly sampling in-batch negatives such as $\langle q_i, d_j \rangle_{i\neq j}$,  and $\langle q_i, q_j \rangle_{i\neq j}$.
However, this method may introduce semantic noise, including false negatives and low-relevance samples, which can hinder the CL training.
To address this, recent work GistEmbed~\cite{solatorio2024gistembed} proposes to use a small but strong guide model to evaluate semantic similarity within a batch, enabling the selection of more informative negatives.
This guided in-batch negative selection improves training efficiency, reduces noise, and enhances embedding quality with minimal computational overhead.
Further, according to the practice view, we prefer a large batch size in CL for training GPTE~\cite{neelakantan2022text}. 
For instance, E5~\cite{wang2022text} utilizes a batch size of 32,768.
Batch sizes in the tens of thousands far exceed the memory capacity of a single GPU.
To carry out training with such large batch sizes, multi-GPU distributed training is necessary—the more GPUs, the larger the batch size you can run.
However, due to resource constraints, one cannot always expand the number of GPUs, which makes performance optimization techniques to increase the batch size on a single GPU essential.
Common techniques include gradient checkpointing (also known as activation checkpointing~\cite{chen2016training}) and the zero redundancy optimizer~\cite{rajbhandari2020zero} of DeepSpeed~\footnote {\url{https://github.com/microsoft/DeepSpeed}}.
Sometimes one cannot use gradient accumulation, while GradCache~\cite{gao-etal-2021-scaling} provides similar functionality.
It does this by extracting and accumulating all embeddings in sub-batches and recomputing gradients, which delivers the same loss and gradient updates.
Although GradCache can execute contrastive learning with very large batch sizes on a limited number of GPUs, it is not generally recommended as a first resort and only explored by a few models~\cite{nussbaum2024nomic}.
This is because it introduces at least double the additional model forward computation, making it a less ideal choice in a pre-training context where efficiency is key.
\subsection{Data Synthesis} \label{sec:role:data}
The rise of PLMs has made data synthesis and augmentation increasingly popular for task-specific optimization, which is further propelled by LLMs. 
GPTE has also benefited from this line~\cite{DBLP:conf/emnlp/0001TSXGJ22,DBLP:conf/emnlp/0001VKC21,DBLP:conf/sigir/WangGZY22,DBLP:conf/emnlp/WangL22,DBLP:conf/emnlp/0001TSXGJ22}, with numerous studies focusing on generating synthetic positive text pairs and designing specialized negative examples to enhance CL training. 

\subsubsection{Training Data Synthesis}\label{sec:train:synth}
Developing text embedding models often necessitates vast amounts of high-quality training data, particularly for supervised fine-tuning.
However, acquiring large-scale, labeled datasets can be expensive and time-consuming.
To address this issue, many works~\cite{lee2024gecko,lee2024nv,wang2023improving,hu2025kalm,qwen3embedding} resort to LLMs to synthesize training data.
Table~\ref{tab:syndata} shows the representative training datasets by LLM synthesis. 
Generally, each training entry for text embedding consists of three parts: anchor, positive, and negative.
The anchor serves as the query, the positive text is semantically relevant or similar to the anchor, and the negative text is semantically irrelevant or dissimilar to the anchor.
Next, we discuss how LLMs synthesize data from these three parts.

Compared with traditional data construction, utilizing LLM to synthesize data is less costly and ensures data quality and diversity, where LLM can generate data for specific tasks, languages, and difficulties according to instructions.
E5-Mistral~\cite{wang2022text} uses a two-stage prompt: the first stage uses GPT-4 to generate diverse task descriptions (such as short text-long text matching), and the second stage generates specific samples including queries, positives, and hard negatives.
Qwen3-embedding~\cite{qwen3embedding} uses Qwen3-32B to synthesize 150 million pairs covering query types, difficulty, and 119 natural and programming languages for pre-fine-tuning, and filters out 12 million high-quality data pairs for fine-tuning.
Gemini Embedding~\cite{DBLP:journals/corr/abs-2503-07891} employs Gemini to filter low-quality examples, determine relevant positive and negative passages for retrieval, and generate rich synthetic datasets.
LLM-synthesized data has become the core driving force behind text embedding model training.

\begin{table*}[t]
\centering
\footnotesize
\tabcolsep=0.45cm
\caption{Representative studies of GPTE data synthesis. We only include studies that regard LLM data synthesis as one of their primary contributions.}
\def\arraystretch{1.1}
\begin{tabular}{ccc}
\hline
\textbf{Category} & \textbf{Methods} & \textbf{LLMs}  \\
\hline
\multirow{7}{*}{\shortstack[c]{ Text Pairs ($x, y^+/y^-$)}  } 
  & Promptagator \cite{DBLP:conf/iclr/DaiZMLNLBGHC23} & FLAN-T5  \\
  & LLM4DE\cite{ma2023pre} & Alpaca/TK-instruct \\
  & SKICSE \cite{DBLP:conf/naacl/OuX24} & LLaMA2 \\
   & SPTAR \cite{peng2025soft} & LLaMA/Vicuna \\
   & Denosent \cite{wang2024denosent} & ChatGPT \\
   & CLAIF  \cite{cheng2023improving} &  GPT3 \\
  & Qwen3-Embedding~\cite{qwen3embedding} & Qwen3\\
\hline
\multirow{8}{*}{\shortstack[c]{ Text Triples ($x, y^+, y^-$)}  }
    & SumCSE~\cite{DBLP:conf/naacl/ThirukovalluruWCLLJD24} & Vicuna \\
    & AdaptCL~\cite{DBLP:conf/coling/0023WWDW24} &  WizardLM\\
    & GenSE~\cite{DBLP:conf/emnlp/ChenZWL022}  & T5 \\
    & SynCSE~\cite{DBLP:conf/emnlp/ZhangLH23} & ChatGPT/GPT4 \\
    & MultiCSR~\cite{DBLP:journals/corr/abs-2310-10962} & ChatGPT/FLAN-T5 \\
    & NLI-GEN~\cite{DBLP:conf/acl/SatoTST24}  & LLaMA2 \\
    & InPars~\cite{DBLP:conf/sigir/BonifacioAFN22} & ChatGPT \\
    & InPars-v2~\cite{DBLP:journals/corr/abs-2301-01820} &  GPT-J \\
\hline
\multirow{4}{*}{\shortstack[c]{ Instructed Triples ($I \circ x, y^+, y^-$)}  } 
    & I3~\cite{DBLP:conf/sigir/Pan0W0S0LLCT24} & ChatGPT  \\
    & Promptriever~\cite{DBLP:conf/iclr/DaiZMLNLBGHC23} & LLaMA3/GPT4o \\
    & Gecko~\cite{lee2024gecko} & N/A \\
    & E5$_{\mathrm{mistral}}$~\cite{wang2023improving} & ChatGPT/GPT4 \\
    & FollowIR~\cite{weller2024followir} & ChatGPT \\
    \hline
\end{tabular}
\label{tab:syndata}
\end{table*}

\textbf{Anchor Text.} The anchor text serves as the initial query from which a text embedding model learns to identify positive and negative text. 
When synthesizing data using LLMs, anchors are typically generated to diversify the input space and create challenging queries.
Initial attempts~\cite{lee2024gecko,lee2024nv,wang2023improving} primarily involve prompting LLMs to generate anchors based on the given document or conditions, \eg specific task types.
Furthermore, some works~\cite{hu2025kalm,qwen3embedding} incorporate `persona' and `difficulty' into generation to increase the diversity and difficulty of the generated data.
In short, these synthesized anchors expand the semantic coverage and improve the robustness of learned text embeddings.
\textbf{Positive Text.} Positive samples can be semantically relevant and similar to the anchor. 
The semantically similar text primarily lies in STS, where the anchor and positive text are symmetric.
LLMs are extensively employed to synthesize semantically similar texts by setting conditions~\cite{DBLP:conf/emnlp/ZhangLH23}, enriching text~\cite{DBLP:conf/naacl/OuX24}, summarizing~\cite{DBLP:conf/naacl/ThirukovalluruWCLLJD24}, etc. 
Conversely, the semantically relevant text is usually asymmetric, where the anchor and positive text are the query and document, respectively.
Here, LLMs play a role in generating either queries from documents or generating both of them simultaneously. 
Many works~\cite{DBLP:conf/iclr/DaiZMLNLBGHC23,DBLP:conf/sigir/BonifacioAFN22,ma2023pre} leverage LLMs' capacity to generate queries from vast document collections. 
Beyond this, some works~\cite{wang2023improving,DBLP:conf/sigir/Pan0W0S0LLCT24} generate both queries and documents based on specific instructions.
\textbf{Negative Text.} 
Negative samples play a pivotal role in CL training. 
Initially, soft negatives, generated through feature transformations such as representation mixing, utilizing lower-layer representations or clustering, are considered effective for generating a large number of negatives~\cite{DBLP:conf/emnlp/GaoYC21,kim2021self,zhang2022virtual,cao2022exploring,zhou2022debiased,zhang2022unsupervised,DBLP:conf/acl/DengWYQW23,chen2023alleviating}\footnote{Several studies explore the use of augmented positive samples; however, their underlying strategy is functionally equivalent to that of augmented negative sampling.}.
Recently, with the remarkable generative capabilities of LLMs, synthetic hard negatives have gained growing interest. 
These hard negatives refer to query-passage pairs that are semantically similar yet practically mismatched.
In the fine-tuning stage, deliberately mining and incorporating such challenging negative examples is key to boosting the GPTE performance. 
LLM-based methods for synthesizing hard negatives generally fall into two categories, based on the application tasks of GPTE.
First, hard negatives for symmetric tasks, which aim to generate contradictory texts to the anchor~\cite{DBLP:conf/acl/SatoTST24,DBLP:conf/aaai/LiZNM24,DBLP:conf/coling/0023WWDW24}, prompt GPTE models to capture subtle semantic distinctions beyond surface-level features. 
Second, hard negatives for asymmetric tasks, which focus on generating contextually irrelevant texts based on the specific definitions of irrelevance to the anchor~\cite{DBLP:conf/iclr/WellerDLPZH25,DBLP:conf/naacl/ThirukovalluruWCLLJD24,DBLP:conf/emnlp/ZhangLH23}, enhance the GPTE models' capabilities in deep semantic understanding. 
\subsubsection{Evolving Benchmark}\label{sec:evolv:bench}
As the application scenarios of embedding models expand and application requirements increase, the benchmarks also need to be expanded in terms of task, language, and text length.
Traditional benchmarks mainly rely on manual annotation and suffer from a static nature, a narrow scope, and a high cost, which makes it difficult to construct or expand benchmarks.
To build a unified assessment task format and fill in gaps in multilingual or task types, using LLMs to synthesize data is a cost-effective method~\cite{zhang2023language}.

For example, in low-resource languages or uncommon task types such as dialogue embedding, LLM is often used to synthesize input pairs, questions, or labels.
In semantic text similarity (STS) or question-answering tasks, LLM is also used to automatically convert formats (\eg generating hypothetical questions or sentence pairs) to maintain consistency in task formats.
Some classification tasks (especially multilingual ones) are constructed using data samples or label explanations automatically generated by LLM.
The AIR-bench~\cite{chen2024air} leverages LLMs to dynamically generate retrieval evaluation data for diverse tasks, domains, and languages.

\begin{figure*}
    \centering
    \includegraphics[width=0.7\linewidth]{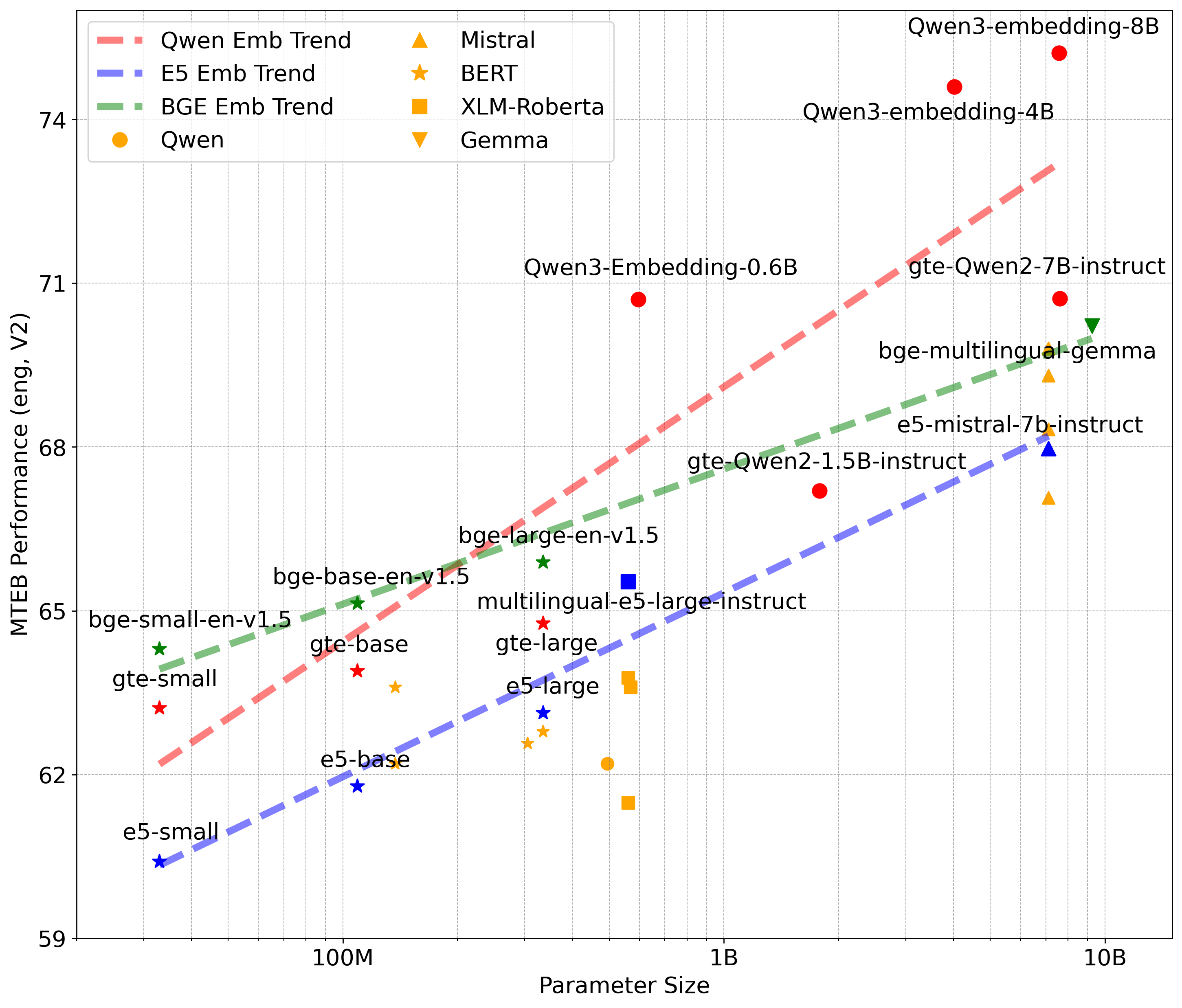}
    \caption{Comparisons of GPTE models with various PLM backbones, focusing on those of widely-adopted open-source PLMs.}
    \label{fig:trend_perf_emb}
\end{figure*}

\subsection{The Choice of PLMs}\label{sec:role:choice}

As the foundations of GPTE, the choice of PLMs is crucial and is influenced by the performance, efficiency, and other advanced capabilities (e.g., multilingualism).
Consequently, this choice is made across two key factors: the model architecture and scale.
Figure \ref{fig:trend_perf_emb} depicts the overall trend of the two factors.

\subsubsection{Comparison of Model Architectures}\label{sec:comp:arch}
The architectural design of PLMs largely impacts their suitability for text embedding.
As summarized in Table \ref{tab:embed-models}, the prevailing architectures for GPTE are encoder-only and decoder-only PLMs. 
Encoder-only GPTE models have dominated the field for several years~\cite{muennighoff-etal-2023-mteb,DBLP:conf/coling/YanoFFTW24}, with BERT serving as the most widely adopted backbone~\cite{DBLP:conf/emnlp/ReimersG19,wang2022text,li2023towards,xiao2023c}, alongside alternatives like XLM-R and RoBERTa. 
Several works also train BERT-alike models from scratch for specific goals such as long-context modeling, e.g.,  GTE-MLM, JinaBERT, and NomicBERT.  
For encoder-decoder PLMs, T5 is the primary choice for GPTE models. 
While most of them use only the encoder component, ST5 explores the decoder as well and finds that decoder-based embeddings perform slightly worse than their encoder-based counterparts.
With the rapid advancement of LLMs, decoder-only models have recently emerged as a popular choice for developing state-of-the-art GPTE models~\cite{zhang2023language,wang2023improving}.
The commonly used decoder backbones include the Qwen series and Mistral, along with others such as Bloom, Gemma, and MiniCPM for training GPTEs.
The widely-accepted decoder backbones to train GPTE models are Qwen series and Mistral, while Bloom, Gemma, MiniCPM and etc. are also used.
Figure~\ref{fig:trend_perf_emb} shows the average performance of various GPTE models, which are collected from the MTEB leadboard.
As a whole, decoder-based models can achieve top-tier performance with sufficient scale and proper training~\cite{zhang2023language,wang2023improving,lee2024nv,li2023towards}.
However, a key observation is that these models often require a significantly larger number of parameters in order to achieve competitive results compared with the encoder-only models.

\subsubsection{Impact of Model Scale}\label{sec:imp:scale}
Larger models with enormous parameters and extensive pre-training on vast corpora tend to possess richer semantic understanding and stronger language modeling capabilities.
Consequently, when pre-trained and fine-tuned for GPTE models, increasing the model scale also leads to more powerful text embedding generally, \textit{e.g.,} gte-Qwen2-7B-instruct \textit{vs.} gte-Qwen2-1.5B-instruct~\cite{li2023towards}.
As shown in Figure~\ref{fig:trend_perf_emb}, the scaling effect has demonstrated that model capacity largely determines the quality of the resulting embedding model.
However, larger models are slower for inference, require more memory, and are more expensive for pre-training and fine-tuning, which inflicts a heavy blow on online applications.
To address this issue, Mixture-of-Experts (MoE) models have emerged as a promising direction for efficient scaling.
Models like GritLM~\cite{muennighoff2024generative} and Nomic-Embed-MoE~\cite{nussbaum2025nomicmoe} employ a sparse MoE architecture, where only a fraction of the model's total parameters (the ``experts'') are activated for any given input text during inference.
This allows them to maintain the representational power of large dense models while achieving the computational efficiency of smaller ones, offering a scalable and cost-effective pathway to scale up high-quality GPTE models without an explosion in inference costs.
More intriguingly, MoEE~\cite{DBLP:conf/iclr/LiZ25} finds that the routing weights used to decide which experts to activate encode meaningful semantic information complementary to the hidden states commonly used as embeddings.
By combining these two sources, MoEE achieves more robust embeddings without any additional finetuning, highlighting that MoE LLMs can serve not only as efficient generators but also as versatile and semantically rich embedding models.

\section{Advanced Roles of PLMs}
\label{sec:advance}
In the previous section, we introduced the basic roles of GPTE in conjunction with PLMs, focusing on standard extensions of text embeddings during the period when PLMs dominated the NLP landscape. 
In this section, we explore several advanced topics that have emerged with the rise of PLMs, 
including multilingual processing, multimodal learning, programming languages, safety considerations, and task-specific adaptation. 
These areas have seen significant progress due to the advent of PLMs, some of which were scarcely explored or even nonexistent prior to their development.    

\begin{table*}[t]
\centering
\tabcolsep=0.35cm
\def\arraystretch{1.1}
\footnotesize
\caption{Representative multilingual GPTE models, where the LLM-based models are not listed as their multilingualism is almost a standard property.}
\begin{tabular}{lccc}
\toprule
 \textbf{Model} &  \textbf{Base Model} &  \textbf{Param Size}  &   \textbf{Language Number} \\
\hline
    GTR \cite{DBLP:conf/emnlp/Ni0LDAMZLHCY22} & mT5  & N/A &  101 \\
    LaBSE \cite{feng2022language} & Transformer (from scratch)  & 471M &  109 \\
    mE5 \cite{wang2024multilingual}  &  XLM-RoBERTa &  118M$\sim$560M & 93 \\
    mContriever \cite{DBLP:journals/tmlr/IzacardCHRBJG22}  & mBERT  & 125M & 104 \\
    mDPR \cite{DBLP:journals/tois/ZhangOML24} &  mBERT & 125M  & 104 \\
    m-ST5 \cite{DBLP:conf/coling/YanoFFTW24}  &  mT5-Enc &  564M$\sim$5.7B  & 101 \\
    mGTE \cite{DBLP:conf/emnlp/ZhangZLXDTLYXHZ24} & Transformer (from scratch)  & 304M & 75 \\
    NLLB-E5 \cite{DBLP:journals/corr/abs-2409-05401} & NLLB  & 600M$\sim$3.3B &  200+ \\
    BGE-M3 \cite{chen2024bge} & XLM-RoBERTa  & 560M &  194 \\
\bottomrule
\end{tabular}
\label{tab:multiemb}
\end{table*} 

\subsection{Multilingualism}\label{sec:multilingual}
Multilingualism has become a central topic in NLP since the introduction of multilingual word representations \cite{mikolov2013exploiting,xing-etal-2015-normalized}.
The development of multilingual PLMs has significantly advanced the ability to GPTE models across languages. 
By adopting multilingual PLMs such as mBERT, XLM, and XLM-RoBERTa as the backbone, we can obtain multilingual text embeddings naturally.
Numerous studies have demonstrated that extending the pretraining and pre-finetuning corpora to include multilingual unlabeled data or machine-translated corpora leads to substantial performance improvements for multilingual text embedding. 
With the encoder-based PLMs, LaBSE~\cite{feng2022language}, mContriever~\cite{DBLP:journals/tmlr/IzacardCHRBJG22}, mDPR~\cite{DBLP:journals/tois/ZhangOML24}, mE5~\cite{wang2024multilingual}, m-ST5~\cite{DBLP:conf/coling/YanoFFTW24}, mGTE~\cite{DBLP:conf/emnlp/ZhangZLXDTLYXHZ24}, NLLB-E5~\cite{DBLP:journals/corr/abs-2409-05401}, BGE-M3~\cite{DBLP:journals/corr/abs-2402-03216}, and GTR~\cite{DBLP:conf/emnlp/Ni0LDAMZLHCY22}, have gained great successes.
Table~\ref{tab:multiemb} provides a summary of several representative multilingual text embedding models along with their corresponding language coverage, offering an overview of the growing ecosystem of multilingual GPTE models.  
More recently, decoder-based LLMs have shown remarkable potential in GPTE.
Since nearly all LLMs inherently support multiple languages with equally impressive effectiveness, we will not address their multilingual capabilities here.
The pretraining of multilingual GPTE heavily depends on the breadth and quality of multilingual corpora used in CL.
Table~\ref{tab:multidata} shows the representative datasets for multilingual GPTE training.
First, a straightforward yet effective approach is to aggregate monolingual corpora across various languages (e.g., English-English and Chinese-Chinese text pairs) to construct a multilingual training set. 
This strategy enables GPTE to capture rich intra-lingual semantics while laying a foundation for robust cross-lingual generalization.
To support this, diverse multilingual resources, e.g., Reddit~\cite{DBLP:journals/corr/abs-2104-07081,sentence-transformers-reddit}, Stackexchange~\cite{DBLP:journals/corr/abs-2104-07081,sentence-transformers-stackexchange}, Wikipedia~\cite{wikidump}, mC4~\cite{xue2021mt5}, CCNet~\cite{DBLP:conf/lrec/WenzekLCCGJG20}, and  xP3~\cite{DBLP:conf/acl/MuennighoffWSRB23},
can be utilized to mine large-scale text pairs. 
Then standard bi-text mining, such as title-body, title-abstract, instruction-output extraction, is applied to curate a multilingual corpus, comprising nearly one billion text pairs across more than 200 languages.
Such scale and diversity provide the semantic coverage and linguistic variation necessary to enhance multilingual text embeddings that generalize well across both tasks and languages.
In addition to the aforementioned data sources, the multilingual versions of high-quality, manually-crafted text pairs from specific tasks can also be leveraged to enhance GPTE training.
These tasks are closely aligned with the GPTE objective, and several have already been adopted as benchmarks for evaluating text embedding models. 
Notable examples include MFAQ~\cite{DBLP:journals/corr/abs-2109-12870}, MLQA~\cite{DBLP:conf/acl/LewisORRS20} and MKQA~\cite{DBLP:journals/tacl/LongpreLD21} for multilingual question-answering, MLSUM~\cite{DBLP:conf/emnlp/ScialomDLPS20}, XL-Sum~\cite{DBLP:conf/acl/HasanBIMLKRS21} for multilingual text summarization, and XNLI~\cite{DBLP:conf/emnlp/ConneauRLWBSS18} for multilingual NLI.
For multilingual text retrieval,  datasets such as MIRACL~\cite{DBLP:journals/tacl/0018TOKAL0RL23} and Mr. TyDi~\cite{DBLP:journals/corr/abs-2108-08787} are widely used.
Moreover, cross-lingual text pairs originally designed for machine translation or cross-lingual IR, 
such as Europarl~\cite{DBLP:conf/lrec/Tiedemann12}, NLLB~\cite{DBLP:journals/corr/abs-2207-04672}, CCMatrix~\cite{DBLP:conf/acl/SchwenkWEGJF20} and CLIRMatrix~\cite{DBLP:conf/emnlp/SunD20},
can serve as valuable resources for learning aligned text representations across languages.

\begin{table*}[t]
\centering
\tabcolsep=3pt
\def\arraystretch{1.1}
\footnotesize
\caption{Representative datasets for multilingual GPTE training. }
\begin{tabularx}{0.99\textwidth}{c|X}
\toprule
 \textbf{Type} & \multicolumn{1}{c}{\bf Datasets} \\ \hline
  \multirow{1}{*}{SS} & PAWS-X~\cite{DBLP:conf/emnlp/YangZTB19}, Europarl \cite{DBLP:conf/lrec/Tiedemann12}, CCMatrix \cite{DBLP:conf/acl/SchwenkWEGJF20}, NLLB~\cite{team2022language,DBLP:conf/emnlp/HeffernanCS22},XNLI \cite{DBLP:conf/emnlp/ConneauRLWBSS18}, XL-Sum~\cite{DBLP:conf/acl/HasanBIMLKRS21}, MLSUM \cite{DBLP:conf/emnlp/ScialomDLPS20}  \\
  \hline
 \multirow{3}{*}{SR} & Reddit, Stackexchange, Wikipedia, Multilingual CC News, mC4 \cite{xue2021mt5}, CLIRMatrix \cite{DBLP:conf/emnlp/SunD20}, CCNet \cite{DBLP:conf/lrec/WenzekLCCGJG20}, Amazon Reviews~\cite{DBLP:journals/corr/abs-2403-03952}, mMARCO~\cite{bonifacio2022mmarcomultilingualversionms}, MLDR~\cite{DBLP:journals/corr/abs-2402-03216}, MIRACL~\cite{DBLP:journals/tacl/0018TOKAL0RL23}, Mr.TyDi~\cite{DBLP:journals/corr/abs-2108-08787}, SWIM-IR~\cite{DBLP:conf/naacl/ThakurNAWLC24}, MFAQ \cite{DBLP:journals/corr/abs-2109-12870}, MLQA \cite{DBLP:conf/acl/LewisORRS20},  MKQA \cite{DBLP:journals/tacl/LongpreLD21}  \\
  \hline 
  \multirow{2}{*}{SE} & AmazonCounterfactual~\cite{DBLP:conf/emnlp/ONeillRKKB21}, AmazonMassiveIntent~\cite{DBLP:conf/acl/FitzGeraldHPMRS23}, AmazonMassiveScenario~\cite{DBLP:conf/acl/FitzGeraldHPMRS23}, MultilingualSentiment~\cite{mollanorozy2023cross}, MTOPIntent~\cite{DBLP:conf/eacl/LiACGGM21}, MTOPDomaint~\cite{DBLP:conf/eacl/LiACGGM21} \\ \hline
  \multirow{1}{*}{Mixed} & xP3 \cite{DBLP:conf/acl/MuennighoffWSRB23}, Aya Dataset~\cite{singh2024ayadatasetopenaccesscollection}, MMTEB~\cite{enevoldsen2025mmteb}  \\ \hline
\end{tabularx}
\label{tab:multidata}
\end{table*} 

LLMs have also emerged as an affordable and scalable solution for generating synthetic data to train multilingual text embedding models.
Recent studies have demonstrated the effectiveness of this approach~\cite{lee2024gecko,qwen3embedding}. 
For example, Swim-IR~\cite{DBLP:conf/naacl/ThakurNAWLC24} introduces SAP (summarize-then-ask prompting), enabling PaLM-2 to generate informative queries in target languages, resulting in a synthetic retrieval training dataset containing 33 languages (high to very-low resource) without any human supervision.
Similarly, JH-POLO~\cite{DBLP:journals/corr/abs-2305-00331} leverages GPT-3 to generate English queries for document pairs in various target languages, producing retrieval data of diverse sizes and linguistic coverage.
mE5-Mistral~\cite{wang2023improving} utilizes GPT-3.5 and GPT-4 to construct training data in 93 languages, facilitating multilingual embedding training at scale. 
While most studies on multilingual GPTE models assume that the multilingual PLM supports the target language and that the target-language CL training data is available, several works investigate less typical conditions.
First, for a target language not covered by the multilingual PLM, the availability of CL data can still facilitate effective GPTE training~\cite{DBLP:journals/tois/ZhangOML24}.
The likely reason is that the CL objective provides strong semantic alignment signals that compensate for the lack of prior language-specific representations.
Another scenario involves the reverse: the PLM supports the target language, but no CL data is available.
In this case, several studies show that leveraging data from other languages can still enhance the quality of target-language embeddings~\cite{DBLP:journals/tois/ZhangOML24,wang2022text,zhang2023language},
highlighting the potential of cross-lingual transfer to mitigate data scarcity.

\begin{table*}[t]
\centering
\tabcolsep=0.45cm
\footnotesize
\caption{Multimodal Embedding, where T, I, and V indicate text, image, and video, respectively.}
\def\arraystretch{1.1}
\begin{tabular}{lccc}
\toprule
   \textbf{Type} &  \textbf{Model} &  \textbf{Base Model} &  \textbf{Modality} \\
\hline
\multirow{7}{*}{\shortstack[c]{ Individual \\ Encoder} }  & CLIP~\cite{DBLP:conf/icml/RadfordKHRGASAM21}  & ViT\&Transformer (from scatch)  & T/I \\
    & BLIP~\cite{DBLP:conf/icml/0008LSH23} & ViT\&BERT  & T/I \\
    & BLIP-2~\cite{DBLP:conf/icml/0001LXH22} & ViT\&OPT/Flan-T5 & T/I \\
    & ALIGN~\cite{DBLP:conf/icml/JiaYXCPPLSLD21}  & EfficientNet\&BERT & T/I \\
    & SigLIP~\cite{DBLP:conf/iccv/ZhaiM0B23}  &ViT\&Transformer (from scratch)& T/I \\
    & SigLIP-2~\cite{DBLP:journals/corr/abs-2502-14786} & ViT\&Transformer (from scratch)  & T/I \\
    & \multirow{1}{*}{Coca~\cite{DBLP:journals/tmlr/YuWVYSW22}} &   Transformer\&Encoder-Decoder & \multirow{1}{*}{T/I} \\ 
    \hline
\multirow{11}{*}{\shortstack[c]{ Unified \\ Encoder} }   & E5-V~\cite{DBLP:journals/corr/abs-2407-12580}  & LLaVA-NeXT & T/I \\
    & VLM2VEC~\cite{DBLP:conf/iclr/JiangMYYZC25}  & Qwen2-VL/LLaVA-NeXT & T/I \\
    & VLM2Vec-V2~\cite{meng2025vlm2vec} & Qwen2-VL & T/I/V \\ 
    & mmE5~\cite{DBLP:journals/corr/abs-2502-08468}& Llama-3.2-Vision &  T/I\\
    & GME~\cite{DBLP:journals/corr/abs-2412-16855}   & Qwen2-VL  & T/I\\
    & BGE-VL~\cite{DBLP:journals/corr/abs-2412-14475}  & LLaVA-NeXT & T/I\\
    & UniME~\cite{DBLP:journals/corr/abs-2504-17432}  & Phi3.5-V/LLaVA-NeXT & T/I\\
    & LLaVE~\cite{DBLP:journals/corr/abs-2503-04812}  & LLaVA-OneVision/Aquila-VL & T/I\\
    & B3~\cite{DBLP:journals/corr/abs-2505-11293}     & Qwen2-VL & T/I\\
    & \textsc{Unite}~\cite{DBLP:journals/corr/abs-2505-19650}     & Qwen2-VL & T/I/V \\
    & Jina-v4~\cite{gunther2025jina}   & Qwen2.5-VL   & T/I/V \\
\bottomrule
\end{tabular}
\label{tab:mmemb}
\end{table*} 

\subsection{Multimodal}\label{sec:multimodal}
The advancement of PLMs has brought powerful linguistic priors and instruction-following capabilities to multimodal embedding.
Table~\ref{tab:mmemb} presents a summary of these representative multimodal embedding models.
Previous works mainly focus on learning multimodal embedding from large-scale, weakly supervised image-text pairs~\cite{DBLP:conf/icml/RadfordKHRGASAM21,DBLP:conf/icml/0001LXH22,DBLP:conf/icml/JiaYXCPPLSLD21,DBLP:conf/iccv/ZhaiM0B23,DBLP:journals/tmlr/YuWVYSW22,DBLP:journals/corr/abs-2502-14786,DBLP:conf/icml/0008LSH23}, where these models usually encode text and images separately and project them into a shared space.
Pioneering breakthroughs like CLIP~\cite{DBLP:conf/icml/RadfordKHRGASAM21} leverage contrastive learning over 400M image-text pairs, training a vision encoder and text encoder to maximize cross-modal alignment via a temperature-scaled cosine similarity objective.
Its successor, BLIP~\cite{DBLP:conf/icml/0008LSH23,DBLP:conf/icml/0001LXH22}, further optimizes efficiency by freezing pre-trained vision encoders and introducing a lightweight querying mechanism for cross-modal interaction.
Subsequent works further advance this paradigm in data curation~\cite{DBLP:conf/icml/JiaYXCPPLSLD21}, loss design~\cite{DBLP:conf/iccv/ZhaiM0B23,DBLP:journals/corr/abs-2502-14786}, training strategies~\cite{DBLP:journals/tmlr/YuWVYSW22}.
Overall, these models share a common paradigm that harnesses massive weakly supervised data for cross-modal contrastive alignment while evolving architectures.
Their success has laid the foundation for integrating PLMs into multimodal embedding, while their disjoint architecture hinders fine-grained fusion, and instruction-following capacity remains limited.
The rapid advancement of PLMs has not only reshaped text embedding but also brought a breakthrough in multimodal embedding.
The integration of PLMs into multimodal embedding has unlocked new potential in fusing diverse modalities and instruction-following capabilities:
(1) Powered by the contextual modeling capabilities of PLMs, multimodal embedding can fuse textual descriptions with visual scenes seamlessly;
(2) Leveraging PLMs' instruction-following capabilities, multimodal embedding can support various tasks, \eg image classification, visual question answering, multimodal retrieval, etc.
Building on these advancements, models like E5-V~\cite{DBLP:journals/corr/abs-2407-12580} fine‐tune on text‐only data yet outperform earlier contrastive methods on image–text retrieval benchmarks by leveraging the PLM's rich semantic priors.
This marks a shift towards leveraging the powerful capabilities of PLMs for more advanced multimodal embedding.

\begin{table}[t]
\centering
\footnotesize
\tabcolsep=0.06cm
\caption{Typical datasets categorized by modalities and tasks, where T, I, VD, and V indicate text, image, visual document, and video, respectively. The single-modal text corpora are also necessary, which is discussed in previous sections.}
\label{tab:mmdata}
\def\arraystretch{1.4}
\begin{tabularx}{0.99\textwidth}{c|c|X}
\hline
\bf Class & \bf Task & \multicolumn{1} {c} {\bf Datasets} \\
\hline

\multirow{1}{*}{Single-Modal} 

  & \multirow{1}{*}{I$\rightarrow$I}  & \multirow{1}{*}{NIGHTS\cite{DBLP:conf/nips/FuTSC0DI23}} \\
\hline
\multirow{4}{*}{ Cross-Modal} 
  & \multirow{2}{*}{T$\leftrightarrow$I} 
  & VisualNews\cite{DBLP:conf/emnlp/LiuWWO21}, Fashion200k\cite{DBLP:conf/iccv/HanWHZZLZD17},
 MSCOCO\cite{DBLP:conf/eccv/LinMBHPRDZ14}, Flickr30k\cite{DBLP:conf/iccv/PlummerWCCHL15}, ImageNet\cite{DBLP:conf/cvpr/DengDSLL009}, 
VisualDial\cite{DBLP:conf/cvpr/DasKGSYMPB17}, Wiki-SS-NQ\cite{DBLP:conf/emnlp/MaL0CL24} \\
\cline{2-3}
  &  \multirow{1}{*}{T$\rightarrow$VD} 
  & TAT-DQA\cite{DBLP:conf/mm/ZhuLFWZC22}, ArxivQA\cite{DBLP:conf/acl/0039WXWFK024}, DocVQA\cite{DBLP:conf/wacv/MathewKJ21}, 
 InfoVQA\cite{DBLP:conf/wacv/MathewBTKVJ22}, VisRAG\cite{DBLP:conf/iclr/YuTXCRYLWHL025}, Colpali\cite{DBLP:conf/iclr/FaysseSWOVHC25} \\
\cline{2-3}
  & \multirow{1}{*}{T$\leftrightarrow$V}
  & MSVD\cite{DBLP:conf/acl/ChenD11}, Tarsier2-Recap\cite{DBLP:journals/corr/abs-2501-07888}, MSR-VTT\cite{DBLP:conf/cvpr/XuMYR16}, 
InternVid-FLT\cite{DBLP:conf/iclr/WangH00YML0C00024}, DiDeMo\cite{DBLP:conf/iccv/HendricksWSSDR17}, CaRe\cite{DBLP:journals/corr/abs-2501-00513} \\
\hline
\multirow{5}{*}{ Fused-Modal } 
  & T$\rightarrow$IT
  & WebQA\cite{DBLP:conf/cvpr/ChangCNGSB22}, EDIS\cite{DBLP:conf/emnlp/LiuFFCW23} \\
\cline{2-3}
  & \multirow{1}{*}{IT$\rightarrow$T} 
  & OVEN\cite{DBLP:conf/iccv/HuLCKJLTC23}, INFOSEEK\cite{DBLP:conf/emnlp/ChenHLSCRC23}, 
 ReMuQ\cite{DBLP:conf/acl/0003FGYB23}, OKVQA\cite{DBLP:conf/cvpr/MarinoRFM19}, LLaVA\cite{DBLP:conf/acl/LinMCB24}, VQA~\cite{antol2015vqa} \\
\cline{2-3}
  & \multirow{1}{*}{IT$\rightarrow$I}
  & FashionIQ\cite{DBLP:conf/cvpr/WuGGARGF21}, CIRR\cite{DBLP:conf/iccv/0002OTG21}, 
 Visual7W\cite{DBLP:conf/cvpr/ZhuGBF16}, RefCOCO\cite{DBLP:conf/emnlp/KazemzadehOMB14} \\
\cline{2-3}
  & IT$\rightarrow$IT
  & OVEN\cite{DBLP:conf/iccv/HuLCKJLTC23}, EVQA\cite{DBLP:conf/iccv/MensinkUCGCZSAF23}, INFOSEEK\cite{DBLP:conf/emnlp/ChenHLSCRC23} \\
\cline{2-3}
  & VT$\rightarrow$V
  & WebVid-CoVR\cite{DBLP:conf/aaai/VenturaYSV24} \\
\hline
\end{tabularx}
\end{table}

In tandem with model advancements, training frameworks and recipes~\cite{DBLP:conf/iclr/JiangMYYZC25,meng2025vlm2vec,DBLP:journals/corr/abs-2504-17432,DBLP:journals/corr/abs-2503-04812,DBLP:journals/corr/abs-2505-11293,DBLP:journals/corr/abs-2505-19650} have evolved beyond standard contrastive learning to enhance discriminative power and instruction-following capabilities.
VLM2VEC-V1/V2~\cite{DBLP:conf/iclr/JiangMYYZC25,meng2025vlm2vec} introduce a method to convert any VLM into an instruction-aware embedding model. 
Others focus on enhancing discriminative power through novel training objectives.
For example, UniME~\cite{DBLP:journals/corr/abs-2504-17432} employs a two-stage approach of knowledge distillation from an LLM teacher and hard negative enhanced instruction tuning, 
while LLaVE~\cite{DBLP:journals/corr/abs-2503-04812} introduces a method to handle hard negatives based on their difficulty dynamically.
Further optimizing the training batch, B3~\cite{DBLP:journals/corr/abs-2505-11293}introduces a novel batch construction strategy that uses a teacher model and community detection to create batches rich in hard negatives; UNITE~\cite{DBLP:journals/corr/abs-2505-19650}develops a universal framework focused on data curation and modality-aware training, introducing modality-aware masked contrastive learning to manage the competitive relationship between different modalities.
A significant challenge in multimodal embedding training remains the scarcity of large-scale, high-quality labeled data. 
To address this, several works~\cite{DBLP:journals/corr/abs-2412-14475,DBLP:journals/corr/abs-2412-16855,DBLP:journals/corr/abs-2502-08468} have investigated data synthesis techniques.
For example, BGE-VL-MLLM~\cite{DBLP:journals/corr/abs-2412-14475} introduces MegaPairs, a method using VLMs to generate a massive synthetic dataset that significantly outperforms models trained on existing data. 
Similarly, GME~\cite{DBLP:journals/corr/abs-2412-16855} develops a pipeline to create  a large-scale, fused-modal dataset specifically for universal multimodal retrieval, where queries and candidates can be any combination of text and images
while mmE5~\cite{DBLP:journals/corr/abs-2502-08468} focuses on the broad scope, cross-modal alignment, and high fidelity for synthesizing effective data to train a powerful multilingual, multimodal embedding model.
Beyond training techniques and data, architectural designs also play a key role.
Jina-v4~\cite{gunther2025jina} introduces a flexible model that supports both single-vector and multi-vector~\cite{DBLP:conf/sigir/KhattabZ20} representations, using task-specific LoRA adapters to optimize performance across diverse scenarios like visually rich image/document retrieval.
In conclusion, current developments span high-quality data synthesis like MegaPairs, improved training techniques like B3's hard negative-rich batch construction, and innovative architecture designs like jina-v4's multi-vector representation. 
It has moved beyond basic cross-modal alignment to building capable, instruction-aware multimodal embeddings that handle diverse tasks, ranging from image, video to visual document tasks.

\subsection{Programming Languages}\label{sec:prog:lang}
Driven by the success of PLMs in NLP, there has been growing interest in applying similar paradigms to programming languages.
Code embedding remains a central focus in the field of code representation learning.
SCELMO~\cite{DBLP:journals/corr/abs-2004-13214} is the first work to introduce code embeddings based on ELMo, 
while CodeBERT~\cite{DBLP:conf/emnlp/FengGTDFGS0LJZ20} and CuBERT~\cite{DBLP:conf/icml/KanadeMBS20} pioneer the use of BERT-style models for code embeddings.
In addition, structural information from programming languages has been leveraged to improve code PLMs.
GraphCodeBERT~\cite{DBLP:conf/iclr/GuoRLFT0ZDSFTDC21} integrates data flow structures to enhance code understanding, 
while models such as SynCoBERT~\cite{wang2021syncobert}, TreeBERT~\cite{DBLP:conf/uai/JiangZLLL21}, and UniXcoder~\cite{DBLP:conf/acl/GuoLDW0022} leverage syntactic structures from abstract syntax trees (ASTs).
DOBF~\cite{DBLP:conf/nips/LachauxRSL21} further combines both data flow and AST information through an implicit strategy, adopting an encoder-decoder framework similar to TransCoder~\cite{roziere2020unsupervised}, PLBART~\cite{DBLP:conf/naacl/AhmadCRC21}, CodeT5~\cite{DBLP:conf/emnlp/0034WJH21}, and CoTexT~\cite{DBLP:journals/corr/abs-2105-08645}. 
These models typically generate code embeddings using the standard GPTE strategy for encoder-based PLMs, 
where the hidden $\langle$CLS$\rangle$ representation is used as the code embedding.

\begin{table*}[t]
\centering
\tabcolsep=0.35cm
\footnotesize
\caption{Representative PLM-based code embedding models, restricted to those that involve CL to drive embedding-focused optimization. }
\def\arraystretch{1.1}
\begin{tabular}{lccc}
\toprule
 \textbf{Model} &  \textbf{Base Model} &  \textbf{Param Size}  &   \textbf{PL Number} \\
\hline
    SynCoBERT \cite{wang2021syncobert}  &  CodeBERT &  125M & 6 \\
    Code-MVP \cite{wang2022code}  & GraphCodeBERT  & 125M & 6 \\
    UniXcoder \cite{DBLP:conf/acl/GuoLDW0022} &  RoBERTa & 125M  & 8 \\
    C3P \cite{DBLP:conf/icdm/PeiLQ022}  &  Transformer (from scratch) &  N/A & 7 \\
    CodeRetriever \cite{DBLP:conf/emnlp/LiGSQZYQJCD22} & GraphCodeBERT  & 125M & 6 \\
    CPT-Code \cite{neelakantan2022text} & GPT3 \& Codex  & 300M$\sim$175B &  N/A \\
    SCodeR \cite{DBLP:conf/emnlp/LiGGLSQJCD22} & UniXcoder  & 125M &  8 \\
    CONCORD \cite{ding2023concord}  & BERT & 125M & 3 \\
    FaCE \cite{DBLP:conf/icassp/LiWZ23a} & RoBERTa  & 125M &  N/A \\
    ContraBERT \cite{DBLP:conf/icse/LiuWXML23} & CodeBERT/GraphCodeBERT  & 125M &  6 \\
    CodeT5+ \cite{DBLP:conf/emnlp/WangLGB0H23} & T5 & 220M$\sim$16B & 9 \\ 
\bottomrule
\end{tabular}
\label{tab:codeemb}
\end{table*} 

CodeGPT\footnote{https://codegpt.co/}, GPT-Neo~\cite{gpt-neo}, GPT-J~\cite{gpt-j}, and Codex~\cite{chen2021evaluating} have sparked a new wave of research in LLM-based code intelligence. 
By fine-tuning GPTs on large-scale code corpora from platforms like GitHub, these models have demonstrated impressive performance on code generation and related tasks. 
Since then, a wide array of LLMs for code have emerged, including AlphaCode~\cite{DBLP:journals/corr/abs-2203-07814}, CodeGen~\cite{DBLP:conf/iclr/NijkampPHTWZSX23}, PaLM-Coder~\cite{DBLP:journals/jmlr/ChowdheryNDBMRBCSGSSTMRBTSPRDHPBAI23}, CodeLLaMA~\cite{DBLP:journals/corr/abs-2308-12950}, DeepSeek-Coder~\cite{DBLP:journals/corr/abs-2401-14196}, CodeQwen~\cite{DBLP:journals/corr/abs-2309-16609}, CodeGemma~\cite{DBLP:journals/corr/abs-2406-11409}, StarCoder~\cite{DBLP:journals/tmlr/LiAZMKMMALCLZZW23}, Phi-1.5/3~\cite{DBLP:journals/corr/abs-2309-05463,DBLP:journals/corr/abs-2404-14219}, Codestral\footnote{https://mistral.ai/news/codestral}, among others. 
The process of deriving code embeddings from these models typically mirrors the standard approach of LLM-based GPTE.
The typical process involves first extracting initial representations through pooling or other advanced techniques, followed by refining these embeddings using CL. 
Despite the modality shift from natural language to programming languages, the underlying methodology for embedding remains fundamentally aligned with that of textual embeddings.
CL has been actively explored to enhance code representation learning, with numerous studies demonstrating its effectiveness~\cite{jain2020contrastive,bui2021self}.
Table~\ref{tab:codeemb} shows the representative code embedding models with CL support. 
Models like UniXcoder~\cite{DBLP:conf/acl/GuoLDW0022}, Code-MVP~\cite{wang2022code},  ContraFlow~\cite{cheng2022path}, TransformCode~\cite{xian2024transformcode} and CONCORD~\cite{ding2023concord} 
exploit code-code pairs for CL pretraining, where these pairs are constructed through various transformation strategies such as feature dropout,  code compression, code clone, identifier renaming, canonicalization, loop exchange, dead code insertion,  compilation-guided multiple-view expansions and etc.
The multi-modal CL with both text and code pairs has also been investigated extensively. 
SynCoBERT~\cite{wang2021syncobert} pioneers the use of text–code and code–code pairs extracted from docstrings, 
while the following representative studies includes C3P~\cite{DBLP:conf/icdm/PeiLQ022}, CodeRetriever~\cite{DBLP:conf/emnlp/LiGSQZYQJCD22}, CPT-Code~\cite{neelakantan2022text}, SCodeR~\cite{DBLP:conf/emnlp/LiGGLSQJCD22},  FaCE~\cite{DBLP:conf/icassp/LiWZ23a}, 
ContraBERT~\cite{DBLP:conf/icse/LiuWXML23} and CodeT5+~\cite{DBLP:conf/emnlp/WangLGB0H23},
exploiting well-mined code–document, code–comment, code-text, and code–code pairs. 

\begin{table*}[t]
\centering
\tabcolsep=3pt
\def\arraystretch{1.1}
\footnotesize
\caption{Representative datasets for training code embeddings, where the reported scales may not be directly comparable due to differences in measurement units.}
\begin{tabularx}{0.99\textwidth}{c|c|c|X}
\toprule
 \textbf{Type}  &  \textbf{Dataset}  &  \textbf{Scale} &  \textbf{Covered Languages}  \\ \hline
      \multirow{3}{*}{Code} & CodeNet \cite{puri2codenet} & 13.9M & 55 languages \\  
           &  Codeparrot (clean)  & 115M &  32 languages \\  
          & StackV2 \cite{DBLP:conf/iclr/SureshRXNMDJ25} & 784.3M &   86 languages  \\  \hline       
    \multirow{8}{*}{Code-Code}  &   \multirow{1}{*}{CodeTrans \cite{lu1codexglue}} &  \multirow{1}{*}{12K} &  Java, C\# \\
          &   \multirow{1}{*}{POJ-104 \cite{mou2016convolutional}} &  \multirow{1}{*}{52K} &  C, C++\\
         &   \multirow{1}{*}{Avatar  \cite{ahmad2021avatar}} &  \multirow{1}{*}{57K} &  Java,Py \\
     &   \multirow{1}{*}{CoST \cite{zhu2022multilingual}} &  \multirow{1}{*}{132K} &  C++, Java, Py, C\#, JS, PHP, C \\
         &   \multirow{1}{*}{CodeTransOcean \cite{yan2023codetransocean}} &  \multirow{1}{*}{270K} &  45 languages \\
          &   \multirow{1}{*}{TransCoder-ST  \cite{roziereleveraging}} &  \multirow{1}{*}{437K} &  Java, C++, Py \\
     &   \multirow{1}{*}{BigCloneBench \cite{svajlenko2014towards}} &  \multirow{1}{*}{1.8M} &  Java\\  \hline       
    \multirow{10}{*}{\shortstack[c]{ Code-Text \\ \& \\ Text-Code}}   &  CoSQA \cite{DBLP:conf/acl/HuangTSG0J0D20}    & 20.6K & Py \\
     &  CONCODE \cite{iyer2018mapping}  &  104K & Java \\
     &  APPS \cite{DBLP:conf/nips/HendrycksBKMAGB21}  & 232K  & Py \\     
      &  CodeFeedback \cite{DBLP:conf/acl/ZhengZSLLFCY24}  &  192K & Py \\
    &  Query4Code \cite{DBLP:conf/emnlp/LiLHZZYLH24}  & 237.2K  & Py \\
    & \multirow{1}{*}{XLCOST \cite{zhu2022xlcost}}  & \multirow{1}{*}{1M}  &  C++, Java, Py, C\#, JS, PHP, C \\
     &  \multirow{1}{*}{CodeSearchNet \cite{DBLP:journals/corr/abs-1909-09436}} &  \multirow{1}{*}{2.3M} &  PHP, Java, Py, Go, JS, Ruby \\
     & \multirow{1}{*}{ CornStack \cite{DBLP:conf/iclr/SureshRXNMDJ25}} & \multirow{1}{*}{ 21.2M} &   PHP, Java, Py, Go, JS, Ruby  \\
     & \multirow{1}{*}{CodeT5+ \cite{DBLP:conf/emnlp/WangLGB0H23}}  & \multirow{1}{*}{37M} & PHP, Java, Py, Go, JS, Ruby, C, C++, C\# \\  
\bottomrule
\end{tabularx}
\label{tab:codedata}
\end{table*} 
Most recently, CODESAGE~\cite{DBLP:conf/iclr/ZhangATDNR0X24} demonstrates the effectiveness of multi-modal CL on large-scale text–function pairs, indicating the scale of training corpora plays a crucial role in code embedding pretraining.
CodeSearchNet~\cite{DBLP:journals/corr/abs-1909-09436} consists of 2.1 million functions paired with their natural language documentation, while CoSQA~\cite{DBLP:conf/acl/HuangTSG0J0D20} offers 20,604 high-quality, manually annotated pairs of natural language queries and code snippets. 
APPS~\cite{DBLP:conf/nips/HendrycksBKMAGB21} provides a diverse set of programming problems and their corresponding solutions, covering various difficulty levels.
More recent datasets continue to expand the scope and depth of text–code interaction. 
CodeFeedback~\cite{DBLP:conf/acl/ZhengZSLLFCY24} introduces 68,000 multi-turn interactions with text-code pairs, 
Query4Code~\cite{DBLP:conf/emnlp/LiLHZZYLH24} constructs 237,200 query–code pairs from 12,300 GitHub repositories, tailored for code retrieval, and CornStack~\cite{DBLP:conf/iclr/SureshRXNMDJ25} presents a high-quality filtered dataset derived from The Stack~\cite{DBLP:journals/tmlr/KocetkovLALMJMF23,DBLP:journals/corr/abs-2402-19173}, 
further enriching the landscape of clean, large-scale code corpora for embedding and retrieval studies.
Table~\ref{tab:codedata} shows the representative datasets that might be beneficial for code embeddings.

\subsection{Adaptation for Specific Scenarios}\label{sec:adapt:spec}
Although GPTE models have achieved impressive performance across a wide range of tasks, there remains considerable room for improvement in scenario-specific settings, such as particular tasks, languages, modalities, domains, or document types.
A straightforward and effective adaptation strategy is to perform SFT using labeled datasets tailored to the target scenario. 
To preserve the strong capabilities of the original GPTE model while adapting to new scenarios, 
lightweight training techniques, such as incorporating neural adapters or applying Low-Rank Adaptation (LoRA), are often used. 
These approaches have proven effective in both the target-scenario performance and the extent of model modification, and are especially popular for small-sized GPTE models 
when BERT-alike or mini-type LLM models are used as backbones.   
Recently, instruction-following embeddings have emerged as a promising direction for task-specific adaptation, as the increasing interest in LLM-based GPTE models~\cite{li2024making}.  
It is a natural extension of the instruct-tuning paradigm for task-specific PLMs from general tasks into text embeddings~\cite{DBLP:conf/nips/Ouyang0JAWMZASR22,DBLP:conf/iclr/SanhWRBSACSRDBX22,DBLP:conf/acl/MishraKBH22,DBLP:conf/iclr/WeiBZGYLDDL22}.
GenSE~\cite{DBLP:conf/emnlp/ChenZWL022} proposes prompt-based CL where prefix prompt templates can be regarded as instructions. 
TART~\cite{DBLP:conf/acl/AsaiSL0I0HY23} introduces the first instruction-following IR system by massive multi-task instruction-tuning. 
INSTRUCTOR~\cite{su-etal-2023-one} annotates instructions for 330 diverse tasks and then trains a GPTE model on this with a contrastive loss, which can achieve good performance on various (new) embedding tasks, including classification, retrieval, and STS. 
LLM2Vec~\cite{behnamghader2024llm2vec} tests the effectiveness of instruction-following text embeddings with LLM-based GPTE models without any supervised training.
INBEDDER~\cite{peng2024answer} further shows that training with question–answer pairs can significantly enhance the instruction-following capability of GPTE models. 
Although multilingual GPTE models have achieved remarkable success, several studies still focus on individual languages. 
Several studies involve training specialized BERT-based models using carefully curated corpora in the target language, from which language-specific text embeddings are derived~\cite{xiao2023c}.
The example attempts include FaBERT~\cite{DBLP:journals/corr/abs-2402-06617} for Persian, NB-BERT for Norwegian, RuBERT for Russian, and GATE~\cite{DBLP:journals/corr/abs-2505-24581} for Arabic.
Only models with reported results on the MTEB leaderboard are listed here. 
In contrast, there are relatively few studies utilizing decoder-based LLMs for language-specific embedding tasks. 
A possible reason is that recent LLMs already demonstrate strong performance across many languages, 
reducing the perceived need for further specialization.
Domain-specific adaptation follows a similar trend to language-specific modeling. 
Most existing studies concentrate on the clinical and biomedical domains. 
Improved text embeddings can be obtained either by using specialized PLMs such as BioBERT~\cite{DBLP:journals/bioinformatics/LeeYKKKSK20} and ClinicalBERT~\cite{DBLP:journals/corr/abs-1904-03323}, or by pretraining on domain-specific contrastive learning corpora, such as DisEmbed~\cite{DBLP:journals/corr/abs-2412-15258}. 
A recent study~\cite{DBLP:journals/corr/abs-2401-01943} shows that generalist models can outperform specialized clinical models on short-context clinical semantic search tasks, potentially reducing the future interest in domain-specific text embeddings.

\section{Future Directions}
\label{sec:future}

In this section, we outline several future directions for GPTE models. 
While for conventional advancements of ongoing and expected progress, such as improved text embedding acquisition strategies~\cite{lee2024nv,muennighoff2022sgpt,gunther2025jina}, enhanced training paradigms (e.g., multi-stage curriculum learning or reinforcement learning)~\cite{qwen3embedding,zhao2025kalmembeddingv2,lee2024nv}, novel optimized objectives~\cite{zhao2025kalmembeddingv2,kusupati2022matryoshka,li2023angle}, better data synthesis methods~\cite{lee2024gecko,qwen3embedding,DBLP:journals/corr/abs-2503-07891}, stronger architectures and backbones~\cite{li2024making,muennighoff2024generative,wang2023improving,qwen3embedding}, as well as the continued development of the advanced roles, we will not extensively elaborate on these established areas. 
Instead, we aim to highlight several future research avenues that are currently underestimated or newly emerging, yet hold significant potential as GPTE models continue to advance.
\subsection{Combination with Text Ranking}\label{sec:comb:rank}
Text ranking is closely related to text embedding, and in many cases, embedding models can be directly applied for ranking tasks~\cite{DBLP:conf/coling/YanoFFTW24}. 
In particular, bi-encoder architectures of GPTE enable efficient deployment by independently encoding query and document representations. 
However, this independent encoding limits the model's ability to capture fine-grained interactions between texts, which are often critical for accurate text ranking. 
As a result, GPTE models may fall short in tasks that demand nuanced relevance estimation. 
In contrast, effective text ranking typically benefits from models that support mutual interactions between query and document, such as cross-encoders, which jointly process both inputs and produce a relevance score directly. 
Therefore, while embedding models offer scalability and versatility, achieving high accuracy in ranking often requires architectures that go beyond the limitations of bi-encoder designs.
Currently, the integration of text ranking into universal GPTE models has received growing attention and is emerging as a promising direction for future development. 
By unifying text embedding and ranking within a single general-purpose framework, the CL datasets can be fully explored, and the strengths of GPTE models can be more effectively leveraged.
Qwen3 Embedding~\cite{qwen3embedding} leverages the instruction-following capabilities of GPTE for text ranking by framing the task as a prompted interaction. 
Specifically, it uses a chat-style prompt that concatenates the query and document, prompting the model to produce a binary output (e.g., "yes" or "no") as an assessment of relevance for ranking purposes.
This is just an initial step toward universalization, which may further advance GPTE beyond embedding and ranking, extending its applicability to a broad range of NLP tasks.
A notable advancement along this line is {Jina-reranker-v3}~\cite{DBLP:journals/corr/abs-2509-25085}, which exemplifies the emerging trend of integrating embedding-based retrieval with interaction-based ranking.
Instead of relying solely on independent document embeddings or late interaction after retrieval, Jina-reranker-v3 introduces a \emph{last-but-not-late} interaction mechanism that applies causal attention between a query and its candidate documents within a shared context window.
This design effectively unifies the strengths of embedding models for efficient large-scale retrieval and reranking models for fine-grained semantic interaction within a single framework.

\subsection{Safety Considerations}\label{sec:safety:consi}
Safety issues have always been a critical research focus in PLMs, and this concern is further amplified with the rapid advancements in LLMs.
While GPTE models built on PLMs demonstrate powerful performance, the security issues they introduce must also be carefully examined.
These risks primarily stem from two aspects: inherent vulnerabilities inherited from upstream pre-trained models, 
and second threats newly introduced or amplified within the embedding models' specific application scenarios.
The pre-training stage of large-scale models often relies on complex and diverse data sources, creating vulnerabilities to data poisoning.
A typical form of such an attack is the backdoor attack, which embeds hidden triggers into the model.
When the input contains this trigger, the model will generate malicious output preset by the attacker.
For instance, Badnl~\cite{DBLP:journals/corr/abs-2006-01043} demonstrates the feasibility of backdoor attacks on large pre-trained models like BERT and highlights their high level of stealth and effectiveness. 
BadCSE~\cite{DBLP:journals/corr/abs-2210-11082} extends this by injecting backdoors through CL, mapping sentences with triggers to the target semantic vector region in the embedding space. 
Further, backdoor injection can also be achieved by modifying the bottom neural network layers without fine-tuning~\cite{DBLP:conf/naacl/YangLZRSH21}.
These works reflect the multiple attack surfaces of embedded models during training, deployment, and distribution.
With the spread use of embedding models, the issue of privacy leakage has become increasingly important.
Attackers can infer some of the original input information from the generated embedding vectors.
For instance, GEIA~\cite{DBLP:conf/acl/0003XS23} trains a generator using generative models (\eg GPT-2 or T5) to reverse embeddings into complete sentences, which achieves a 92\% success rate when inverse 32-token sentences.
ALGEN~\cite{DBLP:journals/corr/abs-2502-11308} uses a small scale (such as 1k) of paired data for alignment, achieving inversion range from different fields and languages.
Beyond reconstructing input texts, inversion attacks can also target the embedding model itself.
Text Revealer~\cite{DBLP:journals/corr/abs-2209-10505}, for example, combines public corpora and GPT-2, using model feedback from the target classifier to guide the iterative optimization of GPT-2's hidden states, thereby reconstructing the private text in the original training set.
Furthermore, surrogate models can be trained to mimic target embedding models, allowing the attacker to infer sensitive information from text embeddings without direct access~\cite{DBLP:conf/acl/HuangTHLL24}.
Overall, the privacy and safety aspects are from the very beginning for text embeddings, which would inevitably gain increasing attention as the growing capabilities of GPTE. 
\subsection{Bias of GPTE}\label{sec:bias:gpte}
The presence of bias in GPTE is an inevitable concern and is expected to emerge as a prominent research focus in the field.
Biases in GPTE are mainly sourced from imbalanced datasets of learning, resulting in task overrepresentation of certain tasks, insufficient linguistic diversity, domain misalignment, and societal stereotypes~\cite{DBLP:conf/nips/BolukbasiCZSK16}. 
These biases not only affect fairness but also limit the robustness, generalizability, and inclusiveness of downstream applications. 
Moving forward, one of the key challenges in embedding research is to develop techniques that identify, mitigate, and evaluate these biases systematically, thereby enabling more reliable and equitable applications.
Task bias arises as current GPTE models are trained on imbalanced task distributions~\cite{DBLP:conf/emnlp/GaoYC21}, with semantic similarity and relevance datasets overrepresented, while feature representation tasks are significantly underrepresented.  
Moreover, the semantic diversity across tasks further amplifies task bias, as different tasks emphasize distinct aspects of meaning.
As a result, the resulting text embeddings tend to overfit to certain tasks, limiting their ability to handle less-represented tasks such as complex text classification, multilabel classification, clustering, and instruction retrieval.
To address this, future research can explore task-agnostic embedding objectives, employ task-diverse pretraining regimes~\cite{qwen3embedding}, and develop meta-task frameworks that promote versatility across a wider range of tasks.
Reducing task bias holds the promise for enhancing generalization, improving performance in zero-shot and compositional scenarios, and increasing flexibility for emerging applications.
Domain, language, and modal biases represent another set of critical concerns. 
Domain bias occurs as GPT models are mostly trained on general corpora, which struggle to perform effectively in specialized areas such as biomedical or scientific texts.
This might be mitigated through domain-adaptive pretraining and in-domain contrastive learning~\cite{DBLP:journals/corr/abs-2504-20595}.
Language bias arises from the dominance of high-resource languages in training data, often resulting in degraded performance for low-resource languages.
Techniques such as multilingual alignment and language-specific tuning can help bridge this gap~\cite{DBLP:conf/emnlp/ZhangZLXDTLYXHZ24,DBLP:journals/corr/abs-2412-08802}. 
Modal bias, particularly in multimodal settings, refers to the over-reliance on textual inputs at the expense of complementary visual, audio, or structured signals.
Balanced modality fusion, modality-invariant representations, and contrastive multimodal learning offer promising paths forward~\cite{DBLP:journals/corr/abs-2404-07983}. 
Addressing these biases can lead to more accurate, inclusive, and versatile embeddings across diverse linguistic, domain-specific, and multimodal contexts.
Finally, social and cultural biases such as gender, race, religion, and nationality would pose serious ethical and societal risks. 
These biases often stem from stereotypes and imbalances present in large-scale training corpora and can propagate into downstream applications~\cite{DBLP:conf/nips/BolukbasiCZSK16}. 
Tackling such biases requires a combination of approaches~\cite{DBLP:conf/acl/LiangLZLSM20}, including counterfactual data augmentation, fairness-aware training objectives, post-hoc debiasing, and the development of robust bias evaluation benchmarks. 
Reducing social and cultural bias not only fosters more ethical and inclusive NLP systems but also builds user trust and aligns AI technologies with broader societal values.
\subsection{Structure Information}\label{sec:struct:info}
One of the most crucial characteristics of text is its inherent structural information. 
While the short-range structural dependencies are largely captured by current GPTE backbones, long-range structure information, which is vital for comprehensive long text understanding, remains significantly under-addressed.
For example, in scientific documents like books, articles, or papers, understanding the overall argument involves connecting concepts and logics from the introduction to the conclusion or linking methodologies to their later results and findings.
In addition to standard texts, structured data such as tables~\cite{DBLP:conf/acl/YinNYR20}, codes~\cite{DBLP:conf/emnlp/FengGTDFGS0LJZ20}, knowledge graphs~\cite{DBLP:journals/tacl/WangGZZLLT21}, and other well-organized information are also essential for effective structural learning.
Without adequately leveraging such structural cues, GPTE models often struggle with tasks requiring global coherence, argumentation flow, or complex multi-hop reasoning, leading to fragmented interpretations rather than holistic understanding.
In fact, structural information has already been explored in the field of information retrieval~\cite{DBLP:journals/corr/abs-2408-08921}.
Representative PLM-based models include the subgraph retriever~\cite{DBLP:conf/acl/ZhangZY000C22}, KG-GPT~\cite{DBLP:conf/emnlp/KimKJC23}, and StructGPT~\cite{DBLP:conf/emnlp/JiangZDYZW23}.
However, all these studies primarily focus on improving fragment embeddings within the structured data, rather than providing a comprehensive embedding of the entire data.
It is desirable to obtain full-data embeddings that, while being decomposable and computationally manageable, enable a global and in-depth understanding of the structured information.
The exploitation of structural characteristics also offers a valuable avenue for task-specific GPTE adaptation. 
For example, a combination of well-organized task instructions and the input text can be naturally conceptualized as a hierarchical discourse graph, inherently forming structured data.
When global and in-depth embeddings can be derived through functional composition with computable operators, we can then easily obtain task-specific text embeddings. 
This would allow GPTE models to better align with the specific demands of diverse tasks by understanding their underlying structural organization.
With support from structure-informed text embeddings, we can advance contrastive learning beyond traditional pairwise texts to encompass chain and graph texts~\cite{DBLP:conf/kdd/FangFZT24}. 
The standard pairwise optimization paradigm, while common, exhibits limitations~\cite{DBLP:conf/emnlp/GaoYC21}. 
For instance, the aforementioned task bias problem can be partially alleviated through fine-grained support during optimization.
As a result, the GPTE model gains a more nuanced understanding by contrasting not just individual text snippets, but their interconnected structural components. 
This allows for a richer learning signal that better captures complex relational patterns and mitigates overfitting to simplistic similarities.
\subsection{Extending GPTE with Reasoning}\label{sec:gpte:reason}
Given the inherent variability in application contexts, GPTE models must be designed with flexibility to accommodate the diverse requirements of downstream tasks and usage scenarios.
The key considerations often include: (1) explainability, ensuring model decisions are interpretable and foster trustworthiness~\cite{DBLP:conf/nips/ZarlengaBCMGDSP22}; 
(2) privacy preservation, particularly crucial in sensitive domains such as healthcare and finance~\cite{DBLP:conf/ht/BeigiSGWL19}; 
and (3) robustness, to maintain consistent performance across noisy, mixed-code, or domain/language-shifted inputs~\cite{qwen3embedding}. 
Consequently, future research should focus on developing embedding models that are not only semantically rich but also inherently adaptable to these evolving practical constraints, thereby enabling safe, effective, and responsible deployment across a wide array of applications.
To meet the above requirements, one promising approach is to design GPTE models that are compatible with a reasoning LLM~\cite{DBLP:journals/corr/abs-2502-07555}. 
By aligning text embeddings with the LLM reasoning mechanisms, it becomes possible to enhance interpretability, support context-aware adaptation, and enable controllable behavior across tasks. 
This compatibility allows embeddings to serve not only as static representations but also as interfaces for dynamic reasoning, enabling applications such as instruction following, multi-step inference, and explainable retrieval. 
Moreover, this integration can facilitate modular system design, where GPTE handles efficient representation learning while the reasoning LLM provides flexible task execution and semantic interpretation. 
Such a hybrid framework could offer a practical path toward embedding systems that are both powerful and responsive to real-world demands.
It is important to note that the above framework is not equivalent to the well-known retrieval-augmented generation (RAG) paradigm, which belongs to one specific application of GPTE. 
Instead, this framework should be viewed as a foundational component of GPTE. 
It enables a shift from abstract, task-agnostic representations to more concrete, context-aware embeddings aligned with reasoning and application needs. 
By coupling GPTE with a reasoning-capable LLM, the embedding space can be made more interpretable, actionable, and adaptable, allowing for richer semantic encoding and more precise downstream task alignment. 
This perspective emphasizes the role of reasoning not merely as an add-on module but as an integral part of embedding generation, bridging the gap between static representation and dynamic understanding.
Recent works such as Think-Then-Embed~\cite{cui2025think} and O1-Embedder~\cite{DBLP:journals/corr/abs-2502-07555} further demonstrate the potential of integrating reasoning mechanisms into embedding generation.
Think-Then-Embed~\cite{cui2025think} introduces a two-stage framework where a reasoning model first produces explicit intermediate thoughts or explanations before the embedder encodes both the original query and the generated reasoning trace.
This paradigm enhances compositional understanding, allowing the embedding model to better capture complex semantic relationships and instruction-level nuances.
Similarly, O1 Embedder~\cite{DBLP:journals/corr/abs-2502-07555} brings reasoning-inspired retrieval into the text embedding, enabling retrievers to ``think before acting.''
By generating internal reasoning signals before retrieval, O1 Embedder achieves substantial gains in multi-task and zero-shot retrieval performance, showing that explicit reasoning can guide embedding models toward more precise and generalizable representations.
Overall, these studies point to a promising evolution of GPTE, where embeddings are not static outcomes of encoding, but dynamic products of deliberate reasoning.
The concept of embedding can be further extended to align with the notion of memory in cognitive neuroscience~\cite{DBLP:journals/corr/abs-2407-01178}, suggesting the potential for high-dimensional, dynamic representations that evolve with ongoing textual, visual, or auditory input. 
Just as human memory encodes, stores, and retrieves information to support reasoning, decision-making, and learning, GPTE serves as a compressed representations that retain essential semantic content for downstream processing. 
In this view, GPTE acts as a form of artificial memory, capable of capturing both episodic and semantic information depending on the task and context. 
By the cognitive perspective, we may get a GPTE model that not only encodes information efficiently but also supports continual learning, contextual adaptation, and memory-augmented reasoning, which provide a more human-aligned foundation for knowledge representation and flexible task execution in complex, real-world environments.

\section{Conclusion}
\label{sec:conclu}
In this survey, we present a systematic review of GPTE in the era of PLMs, highlighting the critical roles that PLMs play in advancing the development of GPTE. 
We begin by introducing the background of GPTE, outlining the fundamental concepts and functions of text embeddings, detailing the general training architecture, and summarizing progress in training data and evaluation benchmarks. 
Next, we examine the foundational roles of PLMs in GPTE, including methods for deriving embeddings, training strategies, diverse learning objectives, and the enrichment of high-quality datasets. 
We also provide a comparative analysis of representative GPTE models with respect to scale and backbone PLMs.
Furthermore, we explore several advanced roles enabled by PLMs, such as support for multilingual and multimodal embeddings, integration with programming languages, and adaptation to diverse real-world scenarios. 
Finally, we discuss promising future directions beyond the current scope of GPTE, including integrating text ranking, addressing safety and bias issues in GPTE, leveraging structural information, and extending GPTE with reasoning capabilities.
We hope this survey serves as a valuable resource for new researchers seeking to understand recent developments in GPTE quickly, as well as for established researchers aiming to comprehensively grasp the current landscape. 
Moreover, we expect that the insights and challenges outlined in this survey will help to guide future research in this rapidly evolving field.

\bibliographystyle{ACM-Reference-Format}
\bibliography{sample-base}

\end{CJK}
\end{document}